\documentclass[11pt]{article}

\usepackage[preprint]{acl}

\usepackage{times}
\usepackage{latexsym}

\usepackage[T1]{fontenc}

\usepackage[utf8]{inputenc}

\usepackage{microtype}

\usepackage{inconsolata}

\usepackage{graphicx}

\usepackage{amsfonts}
\usepackage{hyperref}
\hypersetup{hidelinks}
\usepackage{url}
\usepackage{tabularx}
\usepackage{multicol}
\usepackage{amsmath}
\usepackage[table]{xcolor}
\usepackage{subcaption}
\usepackage{threeparttable}
\usepackage{makecell}
\usepackage{longtable}
\usepackage{placeins}
\usepackage{booktabs}
\usepackage{calc}
\usepackage{siunitx}
\usepackage{float}

\newcommand{\ci}[3][-0.00]{\makebox[\widthof{$\boldsymbol{#1}$}][r]{$#2$}\,$\pm$\,$#3$}
\newcommand{\cib}[3][-0.00]{\makebox[\widthof{$\boldsymbol{#1}$}][r]{$\boldsymbol{#2}$}\,$\pm$\,$#3$}
\newcolumntype{P}[1]{>{\raggedright\arraybackslash}p{#1}}

\title{Dynamic LLM Routers are Often Misguided}

\author{Sam Wang, Julia White, Sahibzada Allahyar, Dhruv Atreja, Urchade Zaratiana, Kelton Zhang \\
Fastino Labs\\
\texttt{\{sam,jwhite,sahibzada,dhruv,uz,kelton\}@fastino.ai}
}

\begin{document}
\maketitle
\begin{abstract}
Dynamic LLM routers promise to cut inference costs by sending each query to the cheapest model that can answer it correctly. We analyze six commercial routers across 14 settings on a diverse benchmark spanning eight task categories, finding that none of them outperforms a router that randomly selects between two well-chosen models at matched cost. Some underperform by more than 10 percentage points. We trace this gap to four patterns prevalent across routers: difficulty blindness, length reversal, semantic matching, and roster suboptimality. We show that the first three are what the standard objective rewards: cost–accuracy Pareto efficiency on realized costs favors escalating moderately hard queries over the hardest ones, shorter queries over longer ones, and routing by a query's source over its difficulty. We also argue that the two assumptions that would justify large rosters, model granularity and model specialization, do not hold empirically. We propose an alternative evaluation methodology that does not reward these patterns, and as a proof of concept, we design a simple two-model router that avoids all four. Nevertheless, its gain over random routing is limited, because a well-chosen roster leaves little to route.
\end{abstract}

\section{Introduction}
\label{sec:intro}
\begin{figure}[t]
    \centering
    \includegraphics[width=\linewidth]{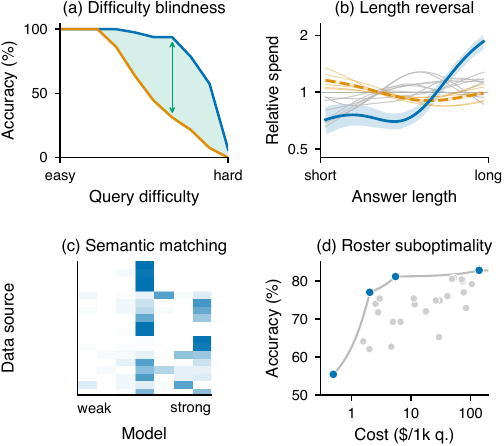}
    \caption{Four patterns of current LLM routers. (a) \emph{Difficulty blindness:} a strong model's (Opus 5 (high)) advantage over a weak one (gpt-oss-20b (low)) peaks at moderately hard queries and shrinks at the expensive end, so routers under-escalate hard queries. (b) \emph{Length reversal:} routing by difficulty (blue) spends more on queries with longer answers, while some commercial routers (orange) spend less. (c) \emph{Semantic matching:} routers often send most of each data source to one model regardless of query difficulty (OpenRouter auto (medium) is shown as a representative example). (d) \emph{Roster suboptimality:} only a small number of models are at the cost-accuracy frontier, yet many commercial routers spread queries across many non-frontier models.}
    \label{fig:panel}
\end{figure}

The central premise of dynamic LLM routing is that the computational expense of inference should scale with query complexity, sending each query to the cheapest model capable of answering it correctly. While some routing architectures attempt to surpass standalone performance by exploiting model specializations \citep{liu2026vllmsemanticroutersignal,tang2026sakanafugutechnicalreport}, the primary goal of routing is cost reduction. We focus on the simplest of these cases: a router for single-turn, general-purpose user traffic that achieves near-frontier performance at a substantial discount.

Despite the intuitive logic of dynamic routing, studies have consistently revealed that learned routers often fail to outperform simple baselines \citep{hu2024routerbench,lu2026routingplateauunderstandingbreaking,li-etal-2026-llmrouterbench}.
Prior work has noted three of the four patterns we study: routers route without regard to query difficulty \citep{lu2026routingplateauunderstandingbreaking} (\emph{difficulty blindness}), collapse each domain onto a single model \citep{li-etal-2026-llmrouterbench} (\emph{semantic matching}), and use rosters that are too large \citep{li-etal-2026-llmrouterbench} (\emph{roster suboptimality}). We identify a fourth, \emph{length reversal}: routers prefer to escalate queries with short answers, even though these tend to be easier.

Our contribution is to explain these patterns. Three of the four are what the conventional evaluation objective rewards. The standard benchmarks \citep{hu2024routerbench,lu2026routerarena,li-etal-2026-llmrouterbench} assess routers for Pareto efficiency on cost and accuracy. However, a router that maximizes this objective escalates a query when an expensive model is much more likely than the cheap one to answer it, and this gap is small on the hardest queries since both models likely fail. A Pareto-efficient router therefore has little reason to escalate the most difficult queries (difficulty blindness). In addition, when answer lengths vary, the same router prefers to escalate queries with shorter expected answers, because escalating them costs less (length reversal). If the evaluation dataset includes many diverse benchmarks of varying difficulty, a router can perform well on evaluations just by predicting the source identity (semantic matching). Answer length and source are also easier to estimate than difficulty, and under this objective they score just as well or better, so routers have little reason to route on difficulty at all. In its stead, we propose an alternative router evaluation methodology that does not reward these behaviors. We attribute the fourth pattern, roster suboptimality, to two assumptions, that intermediate models fill useful price points and that models specialize, and show that neither holds up empirically. We find that a roster of three (and, in most cases, just two) models closely approximates the best achievable accuracy of routers that send harder queries to strong models (\S\ref{sec:roster}).

Finally, we show that a router can avoid all four patterns by directing difficult queries to the stronger model, but on our model set this offers no accuracy gain over random routing at matched cost, because a well-chosen roster leaves little to route.

\section{Related Work}
\label{sec:related}
\paragraph{Dynamic routing.} Many routers train a supervised predictor for a model's answer quality or correctness \citep{ding2024hybrid,lu-etal-2024-routing,chen2024routerdc,feng2025graphrouter,zhuang2025embedllm,varshney2026llmrouterrethinkingrouting,bao2026}, while others route on human pairwise preferences \citep{ong2025routellm}. Semantic routers use BERT-style embedders to classify queries into clusters or categories, assigning each cluster to a single model \citep{jitkrittum2026universal,10.1145/3772429.3772445,zhang2025avengerssimplerecipeuniting,liu2026vllmsemanticroutersignal}.

\paragraph{Router evaluation.} RouterBench \citep{hu2024routerbench}, LLMRouterBench \citep{li-etal-2026-llmrouterbench}, RouterArena \citep{lu2026routerarena}, and TwinRouterBench \citep{yang2026twinrouterbench} all evaluate routers by cost-accuracy Pareto efficiency using realized generation costs. We argue that this objective rewards difficulty blindness, length reversal, and semantic matching.

\paragraph{Router patterns.} RouterBench \citep{hu2024routerbench} found that none of its learned routers outperform a ``zero router'' that randomly interpolates between Pareto-efficient models, which is also the baseline we adopt. \citet{lu2026routingplateauunderstandingbreaking} find that routers learn coarse rather than query-specific signals, and \citet{li-etal-2026-llmrouterbench} that routers do little better than sending each dataset to its best model and that larger rosters show diminishing returns. We find that these patterns persist in commercial routers.

\paragraph{Difficulty and capability.} Item response theory (IRT) describes pass rate as a function of a student (model)'s latent ability and an item (query)'s characteristics \citep{lord1952theory,rasch1960probabilistic,birnbaum1968latent,reckase2009multidimensional}. It has been used to estimate full-benchmark scores from a small subset of items \citep{polo2024tinybenchmarks,kipnis2025metabench}, to place models on a single capability scale \citep{epoch2025arosettastoneforaibenchmarks}, and to predict pass rates for routing \citep{song-etal-2025-irt}. We use the Rasch (one-parameter) model to define model capabilities and query difficulties, and the multidimensional extension to test for specialization. Prior evidence suggests that LLM performance across benchmarks is low- or one-dimensional \citep{burnell2023revealingstructurelanguagemodel,ruan2024observational,ILIC2024101858,epoch2025arosettastoneforaibenchmarks}, and we likewise find specialization too weak to exploit at routing time. Finally, several papers predict difficulty or correctness from model activations \citep{lee2026probingdifficultyperceptionmechanism,cencerrado2026answerneededpredictingllm,varshney2026llmrouterrethinkingrouting,ghasemabadi2026llmspredictfailuresselfawareness,lugoloobi2026llms}, and we build our own difficulty predictor on these methods.

\section{Setup}
\label{sec:setup}
\paragraph{Notation.} Let $\mathcal{M}$ be a \emph{model set}, the set of models we evaluate, and let $\mathcal{Q}$ be a set of queries. For model $i \in \mathcal{M}$ and query $j \in \mathcal{Q}$, $y_{ij} \in \{0,1\}$ is whether the model answers correctly, and $c_{ij}$ is the cost of generating its answer, including both input and output tokens. We define $\overline{c}_{i} = \frac{1}{|\mathcal{Q}|} \sum_j c_{ij}$ as the model's mean cost per query. A \emph{router} $r: \mathcal{Q} \to \mathcal{R}$ assigns each query to a model in its roster $\mathcal{R} \subseteq \mathcal{M}$, with accuracy $\frac{1}{|\mathcal{Q}|} \sum_j y_{r(j),j}$ and cost $\frac{1}{|\mathcal{Q}|} \sum_j c_{r(j),j}$. A \emph{random router} over two models $\{i,k\}$ selects $i$ with probability $p$ and $k$ otherwise, so its expected accuracy and cost are $p$-weighted averages of the two models' values, and varying $p$ traces a line segment between them. We define a query's \emph{difficulty} as $d_j = 1 - \frac{1}{|\mathcal{M}|} \sum_i y_{ij}$, the fraction of models that fail it. A model's \emph{ability} $a_i$ is its Artificial Analysis Intelligence Index score \citep{artificialanalysis2026aaii}, which we only use to order models. A query's \emph{expected answer length} is $\ell_j = \frac{1}{|\mathcal{M}|}\sum_i n_{ij}$, where $n_{ij}$ is the number of output tokens model $i$ generates for query $j$. Every query belongs to a \emph{data source} $g_j$ (Table~\ref{tab:sources}). We also fit a Rasch model with capability $\theta_i$ per model and a difficulty per query. The fitted difficulties induce the same ordering as $d_j$, so with a mild abuse of notation, we write $d_j$ for them as well. Capabilities $\theta_i$ are fit from our outcomes and distinct from $a_i$.

\paragraph{Commercial routers.} We analyze six existing commercial routers across 14 settings. We evaluate commercial rather than academic routers because they can be used off the shelf and presumably draw on the best available techniques. The developers include general platforms (OpenRouter, Microsoft Azure, and vLLM) and routing specialists (Not Diamond, Nadir, and Orca). For each router and query, we record the selected model then independently generate and evaluate its generation using default OpenRouter settings and costs, so all costs and accuracies are measured the same way. Appendix~\ref{app:setup} describes each router, the input format, and several notable omissions because they are not direct comparisons.

\paragraph{Datasets.} Our data comprise 17 benchmarks across eight categories (Table~\ref{tab:sources}), with 2{,}508 training samples and 800 evaluation samples. The categories are coding, instruction-following, knowledge, math, question-answering (QA), office, tool use, and Humanity's Last Exam (HLE) (a challenging benchmark to assess the frontier of model capability), together spanning a wide sweep of LLM use cases. Each category has 100 evaluation samples and contributes equally to evaluation scores. All sources admit verifiable rewards except for GDPval, which provides detailed rubrics for LLM judges, and Humanity's Last Exam, where the judge only has to compare the provided answer with the answer key.

\paragraph{Models and evaluation.} We generate and grade responses across 26 models (\S\ref{app:setup}). We selected 20 of them in early June 2026 to cover the frontier at the time, and added six in mid-August as new models were released, giving us two snapshots of the model frontier. Unless stated otherwise, results use all 26 models. We use Gemini 3.7 Flash and Opus 5 (high) throughout this paper because the former is strong for its price point and the latter has the highest accuracy. To ensure that our observations are not limited to this time frame, we replicate findings on RouterBench \citep{hu2024routerbench} (11 models, 34{,}381 queries), SPROUT \citep{somerstep2025carrotcostawarerate} (13 models, 44{,}241 queries), and LLMRouterBench \citep{li-etal-2026-llmrouterbench} (33 models, 3{,}185 queries), using queries where every model has a graded answer.

\section{Common Router Patterns}
\label{sec:analysis}
\begin{table*}[t]
    \centering
    \small
    \setlength{\tabcolsep}{4pt}
    \caption{Comparison between commercial routers and a random \{Gemini 3.7 Flash, Opus 5 (high)\} router at matched cost (\S\ref{sec:setup}), with our router (\S\ref{sec:router}) below. $\Delta$ is the router's accuracy minus the random router's, with a 95\% paired Wald interval (which is narrower for Nadir because both it and the random baseline route many queries to Opus 5). $\gamma(d,a)$ is Goodman-Kruskal's gamma between query difficulty $d$ and the chosen model's ability $a$ (\S\ref{app:aaii}); $\gamma(\ell,c)$ is between expected answer length and the chosen model's mean cost; \emph{Source AMI} is the adjusted mutual information between source and routing choice, the share of routing decisions explained by the source (\S\ref{app:def}); \emph{Within-source $\gamma(d,a)$} is the mean per-source $\gamma(d,a)$. These intervals are 95\% jackknife intervals over queries. Our router is fit out of distribution (\S\ref{app:setup}), and Table~\ref{tab:ours-id} gives in-distribution values. A bold point estimate indicates a ``pass'': not significantly below zero for $\Delta$ and $\gamma(\ell,c)$, and significantly above zero for $\gamma(d,a)$, globally or within source. AMI is reported for context but not graded.}
    \label{tab:router-patterns}
    \begin{tabular}{lccccc}
    \toprule
    & Accuracy
    & \makecell{Difficulty\\blindness}
    & \makecell{Length\\reversal}
    & \multicolumn{2}{c}{Semantic matching} \\
    \cmidrule(lr){2-2} \cmidrule(lr){3-3} \cmidrule(lr){4-4} \cmidrule(lr){5-6}
    Router
    & $\Delta$ (pp)
    & $\gamma(d,a)$
    & $\gamma(\ell,c)$
    & Source AMI
    & \makecell{Within-source\\$\gamma(d,a)$} \\
    \midrule
    OR auto (med)                     & \ci[-00.0]{-5.5}{2.1} & \ci{+0.08}{0.08} & \ci{-0.24}{0.07} & \ci{0.49}{0.04} & \ci{-0.12}{0.12} \\
    OR auto (high)                    & \ci[-00.0]{-4.0}{2.0} & \ci{-0.16}{0.06} & \cib{+0.02}{0.05} & \ci{0.59}{0.04} & \ci{-0.02}{0.12} \\
    OR auto (xhigh)                   & \ci[-00.0]{-3.3}{2.0} & \ci{-0.01}{0.06} & \cib{+0.08}{0.05} & \ci{0.60}{0.04} & \ci{+0.00}{0.12} \\
    OR auto-$\beta$ (med)             & \ci[-00.0]{-7.7}{2.5} & \ci{+0.00}{0.09} & \ci{-0.10}{0.08} & \ci{0.56}{0.05} & \ci{-0.10}{0.17} \\
    OR auto-$\beta$ (high)            & \ci[-00.0]{-8.8}{2.5} & \ci{-0.07}{0.09} & \ci{-0.16}{0.08} & \ci{0.54}{0.06} & \ci{-0.08}{0.17} \\
    OR auto-$\beta$ (xhigh)           & \ci[-00.0]{-2.0}{1.9} & \ci{-0.03}{0.06} & \cib{-0.01}{0.05} & \ci{0.57}{0.04} & \ci{+0.01}{0.12} \\
    Azure (balanced)                  & \ci[-00.0]{-8.4}{2.6} & \cib{+0.10}{0.07} & \cib{+0.01}{0.06} & \ci{0.36}{0.04} & \ci{+0.10}{0.10} \\
    Azure (quality)                   & \ci[-00.0]{-2.2}{1.9} & \cib{+0.18}{0.07} & \ci{-0.19}{0.06} & \ci{0.37}{0.04} & \ci{-0.08}{0.11} \\
    ND (cost)                         & \ci[-00.0]{-8.8}{2.7} & \ci{+0.07}{0.08} & \cib{+0.29}{0.07} & \ci{0.68}{0.04} & \ci{-0.03}{0.15} \\
    ND (quality)                      & \ci[-00.0]{-4.1}{2.2} & \ci{-0.12}{0.08} & \ci{-0.25}{0.07} & \ci{0.60}{0.03} & \ci{-0.01}{0.13} \\
    vLLM SR                           & \cib[-00.0]{-1.0}{1.8} & \cib{+0.15}{0.06} & \cib{+0.10}{0.05} & \ci{0.79}{0.04} & \ci{+0.11}{0.17} \\
    Nadir (2-tier)                    & \ci[-00.0]{-1.4}{1.1} & \ci{+0.12}{0.14} & \cib{+0.71}{0.07} & \ci{0.45}{0.08} & \ci{+0.21}{0.21} \\
    Nadir (3-tier)                    & \ci[-00.0]{-1.8}{1.2} & \cib{+0.12}{0.11} & \cib{+0.63}{0.07} & \ci{0.39}{0.06} & \cib{+0.17}{0.16} \\
    Orca (adaptive)                   & \ci[-00.0]{-10.5}{2.8} & \ci{+0.05}{0.06} & \cib{+0.04}{0.05} & \ci{-0.01}{0.01} & \ci{+0.06}{0.07} \\
    \midrule
    \rowcolor{gray!15} Ours (\$10/1k q.)       & \cib[-00.0]{-0.2}{0.2}  & \cib{+0.40}{0.18} & \cib{+0.64}{0.14} & \ci{0.36}{0.08}  & \cib{+0.48}{0.23} \\
    \rowcolor{gray!15} Ours (\$20/1k q.)       & \cib[-00.0]{-0.7}{0.7}   & \cib{+0.29}{0.11} & \cib{+0.49}{0.10} & \ci{0.48}{0.08}  & \cib{+0.53}{0.16} \\
    \rowcolor{gray!15} Ours (\$40/1k q.)       & \cib[-00.0]{-0.7}{1.0}  & \cib{+0.20}{0.10} & \cib{+0.39}{0.08} & \ci{0.37}{0.07}  & \cib{+0.29}{0.17} \\
    \bottomrule
    \end{tabular}
\end{table*}

A router should outperform any random routing scheme at matched costs, but we find that a simple router that randomly chooses between \{Gemini 3.7 Flash, Opus 5 (high)\} significantly outperforms every commercial router except for vLLM Semantic Router, which was fit on our training data (Table~\ref{tab:router-patterns}). Even when we restrict the random baseline's roster to models each router had access to, no commercial router can outperform it, so this performance gap is not because we are using newer and more cost-efficient models.\footnote{All routers include the GPT 5.6x series (released July 9, 2026), and some include Gemini 3.7 Flash (released August 13, 2026). For routers without Gemini 3.7 Flash, the random router over \{GPT 5.6 Luna (high), GPT 5.6 Sol (xhigh)\} is at least as strong (Table~\ref{tab:commercial-full}).}  We find that every commercial router that we evaluated exhibits several of the following four patterns:

\paragraph{Difficulty blindness.} Across all routers, routing choices correlate at most weakly with query difficulty, and we argue that the standard objective gives them little reason to. We measure the Goodman-Kruskal gamma (\S\ref{app:def}) between query difficulty $d_j$ and the chosen model's ability $a_i$, and find that no router exceeds $\gamma=+0.20$.

\paragraph{Length reversal.} Queries with longer answers tend to be more difficult, with $\gamma=+0.52$ between answer length and difficulty in our evaluation set, so a router responding to difficulty should send longer-answer queries to stronger models. However, $\gamma$ between a query's expected answer length $\ell_j$ and the chosen model's mean cost $\overline{c}_i$ is significantly negative for five commercial router settings.

\paragraph{Semantic matching.} Queries within a source vary in difficulty, but most commercial routers send nearly all of a source to one model (Figure~\ref{fig:source-map}), which suggests that they are routing by source identity. We measure this in two ways: AMI (adjusted mutual information) between the source identity and routing choices (see \S\ref{app:def}), and mean within-source $\gamma$ between query difficulty $d_j$ and the chosen model's ability $a_i$. vLLM SR, which routes by category, has the highest AMI. Orca achieves essentially zero AMI because it spreads each source over dozens of models.

\paragraph{Roster suboptimality.} A roster should contain the most cost-efficient models and nothing else, and it is suboptimal if it omits one of them or if it includes others, which only add ways to route wrongly. This suggests a simple test: if a router underperforms a random router over two well-chosen models, either its roster or its routing policy must be suboptimal, and most commercial routers fail this test (Table~\ref{tab:router-patterns}). Most rosters include many models that do not help, while Nadir's omits frontier models, routing only among Claude models. Its cheaper tiers contain Haiku 4.5 and Sonnet 5, neither of which is on the frontier.\footnote{At the time of access, Nadir offered Haiku 4.5, Sonnet 5, Opus 5, and Fable 5 for each tier. Any cost-saving configuration must use Haiku or Sonnet as the cheap tier, which are not on the Pareto frontier. Our configurations are the only cost-saving ones the platform allows.}

No router correlates with difficulty above $\gamma=+0.20$, five settings exhibit significant length reversal, most are far more source-concentrated than difficulty-based routing produces (Table~\ref{tab:ours-id}) while being blind to within-source difficulty, and most fail the roster test. The next two sections examine the origins of these patterns.

\section{Rethinking Pareto Efficiency}
\label{sec:eval}
\begin{figure*}[h]
    \centering
    \begin{subfigure}{0.49\linewidth}
        \centering\includegraphics[width=0.85\linewidth]{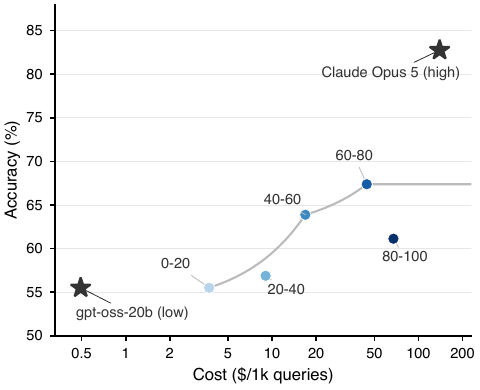}
        \caption{Difficulty-based escalation.}
    \end{subfigure}
    \begin{subfigure}{0.49\linewidth}
        \centering\includegraphics[width=0.85\linewidth]{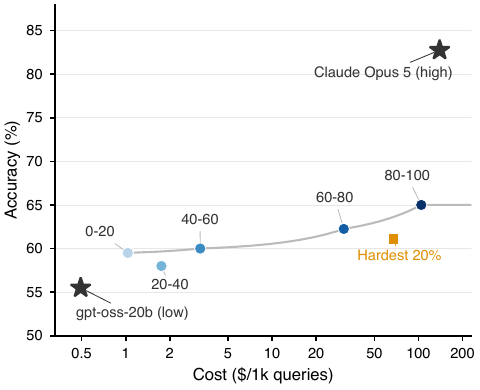}
        \caption{Answer length-based escalation.}
    \end{subfigure}
    \caption{Accuracy against cost when escalating a 20\% band of queries from the cheap to expensive model based on different metrics. (a) escalates based on the difficulty of the query based on model pass rates across the whole model set, while (b) escalates based on output length, also averaged over the model set. In (b), we mark the reference point corresponding to hardest-20\% escalation. 95\% half-widths do not exceed 3.5~pp, and in (a), the 60--80 band outperforms the 80--100 band by $6.3 \pm 3.0$~pp. The same trends hold across other datasets (Figure~\ref{fig:escalation-full}).}
    \label{fig:escalation}
\end{figure*}
The standard router evaluation methodology aims for Pareto efficiency on cost and accuracy. Many routers optimize for this objective directly with a rule of the form $\arg\max_i \hat{q}_{ij} - \lambda \cdot \hat{c}_{ij}$, where $\hat{q}_{ij}$ is model $i$'s predicted accuracy on query $j$, $\hat{c}_{ij}$ is the predicted cost, and the cost penalty $\lambda$ trades off between the two. When the predictions are accurate, this rule is Pareto-optimal for expected accuracy under an expected-cost budget \citep{jitkrittum2023when,dekoninck2025a}. Whether or not commercial routers use this rule, they are likely measured against the same objective. We show that this version of Pareto efficiency encourages some undesirable behaviors.

\subsection{Difficulty blindness}
If only 20\% of queries can be escalated, the natural choice is the hardest 20\%, but this is not Pareto efficient. Among 20\% difficulty bands, escalating the 40--60th or 60--80th percentiles is Pareto-efficient but escalating the 80--100th is not, both on our evaluation set (Figure~\ref{fig:escalation}) and on other datasets (Figure~\ref{fig:escalation-full}). This is because Pareto efficiency focuses on the tradeoff between cost and accuracy, and the strong model surpasses the weak model the most on moderately difficult queries. On the easiest queries, the marginal gain is small because the weak model is likely to succeed, and on the most challenging queries, the gain is also small because both models are likely to fail. Under a Rasch model with capabilities $\theta_1 > \theta_2$, the gain from escalation is $\sigma(\theta_1 - d) - \sigma(\theta_2 - d)$, which peaks at $d=(\theta_1+\theta_2)/2$ and vanishes to 0 as $d \to \pm \infty$.

\subsection{Length reversal}
Pareto efficiency also rewards escalating queries with shorter answers, as each answer contributes equally to accuracy but unequally to cost, and we see this when we escalate instead based on expected answer length (Figure~\ref{fig:escalation}b). On our evaluation dataset, we can achieve 59.5\% accuracy at \$1.03/1k queries by escalating the 0--20th length percentiles, just 1.6~pp below escalating the hardest 20\% at 1.5\% of the price, and outperform the hardest 20\% in accuracy at 46\% of the cost by escalating the 60--80th length percentiles. We call this behavior \emph{accuracy hacking}, in which a router invokes the smarter model when it is cheapest to do so.

This phenomenon explains why we sometimes observe length reversal among commercial routers. The CARROT design \citep{somerstep2025carrotcostawarerate}, in which a router predicts both the model's answer quality and length, directly encourages this behavior by penalizing longer answers from expensive models. A router that uses CARROT with a high cost penalty $\lambda$ will likely exhibit length reversal.

\subsection{Semantic matching}
\begin{figure*}[h]
    \centering
    \begin{subfigure}{0.49\linewidth}
        \centering
        \includegraphics[width=0.8\linewidth]{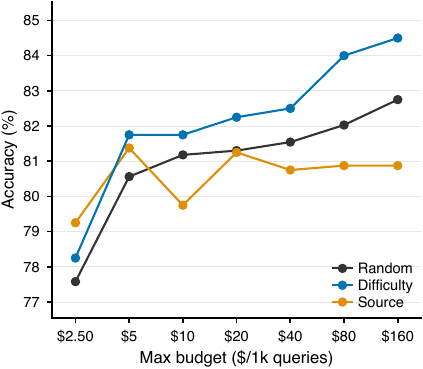}
        \caption{Realized per-query costs.}
    \end{subfigure}
    \begin{subfigure}{0.49\linewidth}
        \centering
        \includegraphics[width=0.8\linewidth]{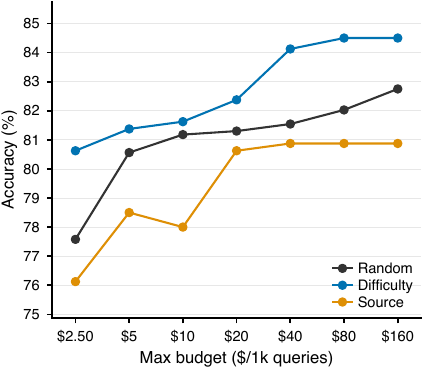}
        \caption{Uniform per-model costs.}
    \end{subfigure}
    \caption{Accuracy at each maximum budget using three routing methods. \emph{Random} is random routing between the two frontier models adjacent to the budget. \emph{Difficulty} is a router with perfect knowledge of query difficulty, choosing its roster of up to five models and its thresholds on the evaluation set itself (\S\ref{sec:roster}). \emph{Source} maps each data source to a single model, chosen on the training set. In (a), each query is charged what the answer actually cost, while in (b), each query is charged that model's average inference cost. Since a map chosen on the training set costs a different amount on the evaluation set, we fit source maps on a dense grid of training budgets and use the highest-budget map that stays under each evaluation budget. Half-widths for the gaps between difficulty and random routing are at most 1.6~pp, and those between difficulty and source routing are at most 2.1~pp.}
    \label{fig:semantic}
\end{figure*}
Semantic matching can look attractive using realized per-query costs, where source routing outperforms random routing at the two cheapest budgets (Figure~\ref{fig:semantic}). But when we use uniform per-model costs, it either underperforms or ties, suggesting that much of semantic matching's apparent advantage at these ranges was in accuracy hacking. Answer length tracks source closely, and source is easier to predict than difficulty, so source routing can look better on this metric than difficulty routing. Semantic matching is particularly brittle because the training dataset's composition determines the router's choices, which might generalize poorly. For example, if the math questions in the training set tend to be challenging, then a router learns to choose a strong model for all math questions, even trivial ones.

\subsection{A revised methodology}
A robust evaluation methodology should reward the routing behavior we want. Two of the three corrections follow from the analysis above, and the first depends on the following judgment. On a query no model can solve, the standard objective is indifferent between cheap and expensive models, since both fail. We take the position that users would prefer if the more capable model attempted it, and would query that model themselves if they knew the weaker model had been used. Partial credit could make this difference measurable, but assigning it is difficult, so we adopt our assumption as a design choice and treat correction 1 as optional for readers who disagree. We propose the following corrections for the patterns:
\begin{enumerate}
    \item \textbf{Difficulty blindness:} Escalation on the most difficult queries should be directly encouraged, not penalized. We achieve this by partitioning our evaluation set into a \emph{hard band} $H \subseteq \mathcal{Q}$ and an \emph{easy band} $E \subseteq \mathcal{Q}$, where $H\,\dot{\cup}\,E=\mathcal{Q}$, and optimizing for Pareto efficiency jointly on three metrics: high hard-band escalation rate, high easy-band accuracy, and low easy-band cost.
    \item \textbf{Length reversal:} A router chooses which model answers but not how long the answer is, so we charge each answer from model $i$ its mean cost $\overline{c}_i$ rather than the realized cost $c_{ij}$, just as every answer contributes equally to accuracy.
    \item \textbf{Semantic matching:} When we predict a model's pass rate, we should evaluate the predictor's mean $\gamma$ within any given data source, not over the entire dataset. Then, a predictor that assigns a single pass rate to each source cannot score well.
\end{enumerate}
Note that we do not assess a router by its accuracy or cost on the hard band since we want it to escalate difficult queries, even impossible ones, to stronger and more expensive models. The size and composition of the hard band is subjective, depending both on the models in the roster and the target price point of the router. In our case, we define the hard band as any query with at least a 50\% failure rate ($H = \{j:d_j \geq 0.5\}$), which encompasses 26.2\% of the evaluation set.

\section{Why Are Rosters Large?}
\label{sec:roster}
Most commercial routers we evaluate route among many models that are not on the cost-accuracy frontier. We believe two appealing but empirically unjustified assumptions explain why.

\subsection{Model granularity}
\begin{figure}[t]
    \centering\includegraphics[width=0.8\linewidth]{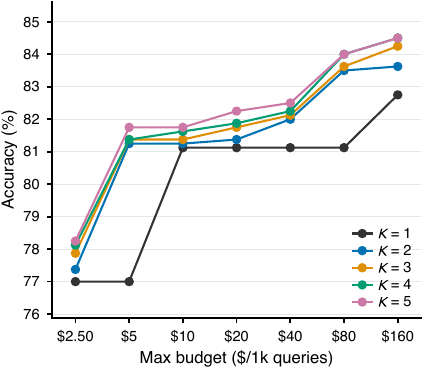}
    \caption{For each budget, the highest accuracy a router can achieve with perfect knowledge of query difficulty, choosing the roster and thresholds on the evaluation set itself, for rosters of $K$ models, where harder queries are sent to more capable models. Accuracies above the strongest standalone model reflect overfitting to the evaluation set, since stronger models typically outperform weaker ones across all difficulty bands (Figure~\ref{fig:rasch}).}
    \label{fig:roster-bloat}
\end{figure}

If we only route between a cheap and an expensive model, then a medium-difficulty query has two poor options: the cheap model, which risks a wrong answer, or the expensive model, which costs far more. We could resolve this with an intermediate model, allowing us to match each difficulty level to the appropriate price level. At the limit, we would have a large roster, where each model is matched to a narrow difficulty range.

Empirically, however, few models appear on the cost-accuracy frontier in any model set, and even fewer are useful for routing (Figures~\ref{fig:paretos} and~\ref{fig:granularity-full}). In the 26-model set, Gemini 3.7 Flash only trails Opus 5 (high) by 1.6~pp while maintaining a substantial 96\% discount. Even when a router knows each query's true difficulty and chooses its roster and thresholds on the evaluation set itself, which is the most favorable setting possible, a roster of two models is never more than 1~pp below the best roster of any size (Figure~\ref{fig:roster-bloat}). Larger rosters are also more fragile, with more thresholds to tune from noisy difficulty estimates, so the small gains they show here, measured in-sample, may not generalize out of distribution.

\subsection{Model specialization}
Even if two models have the same overall capability, they may specialize in different areas. If so, a router should route on domain as well as difficulty, and the roster should include models with different specializations to cover different task types. We test this hypothesis using item response theory (IRT): if models do not specialize, then each can be described by a single capability $\theta_i$ and each query by a difficulty $d_j$, where model $i$ passes query $j$ with probability $\sigma(\theta_i - d_j)$. This is the Rasch model and it fits our outcomes well (Figure~\ref{fig:rasch}). If models do specialize, then a model's propensities $p_i \in \mathbb{R}^k$ interact with a query's features $f_j \in \mathbb{R}^k$ beyond capability and difficulty alone, giving $\sigma(\theta_i - d_j + p_i \cdot f_j)$ \citep[as in][but with separate capability and difficulty parameters]{song-etal-2025-irt}.

To establish an upper bound on this approach, we fit the model parameters $\theta_i$ and $p_i$ on the evaluation set directly. For a given query $j$, the difficulty $d_j$ and features $f_j$ are fit with one model's outcome masked at a time, and we evaluate on the masked model's outcome to prevent overfitting. We also estimate query features from Qwen3.5-27B \citep{qwen3.5} activations, or using a combination of masked pass rates and activations. These settings are highly generous, since a deployed router can only route on query features.

\begin{figure*}[h]
    \centering
    \begin{subfigure}{0.49\linewidth}
        \centering\includegraphics[width=0.7\linewidth]{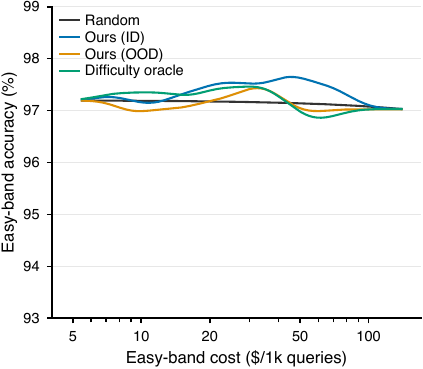}
        \caption{Easy-band accuracy against easy-band cost.}
    \end{subfigure}
    \begin{subfigure}{0.49\linewidth}
        \centering\includegraphics[width=0.7\linewidth]{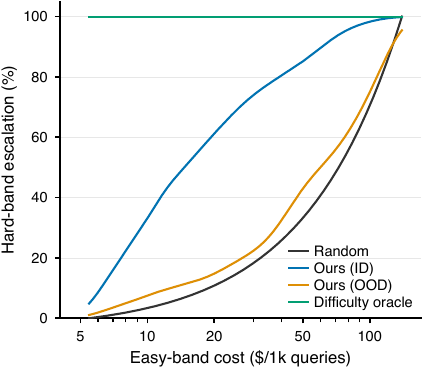}
        \caption{Hard-band escalation against easy-band cost.}
    \end{subfigure}
    \caption{Our router on \{Gemini 3.7 Flash, Opus 5 (high)\} evaluated using our revised methodology from \S\ref{sec:eval}, where all costs are per-model mean costs and the hard band contains queries with at least a 50\% failure rate. Random is random routing between the two models. Ours (ID) and (OOD) are our router fit in and out of distribution (\S\ref{app:setup}). Difficulty oracle uses the true difficulty scores to route. On (a), 95\% confidence interval half-widths for the gap with random do not exceed 1.2~pp, with all deviation from random routing remaining within noise; on (b), they do not exceed 7.5~pp, with our router surpassing random routing on hard-band escalation outside noise across \$6--139/1k queries ID, and \$42--73/1k and \$121--139/1k queries OOD.}
    \label{fig:g37-opus-pair}
\end{figure*}
Specialization effects are weak even in these generous settings, where increasing $k$ from 0 to 8 reduces NLL only from $0.27 \pm 0.02$ to $0.24 \pm 0.02$ (Figure~\ref{fig:mirt}). This is unsurprising, since frontier models are trained as generalists and prior work found that LLM performance across benchmarks is well described by a single capability score \citep{ruan2024observational,ILIC2024101858,epoch2025arosettastoneforaibenchmarks}. Specialization can still matter when routing among frontier models \citep{tang2026sakanafugutechnicalreport}, or on older, more heterogeneous model sets such as SPROUT, where source-based routing outperforms difficulty-based routing (Figure~\ref{fig:semantic-all}e--f). But in a roster of cheap and expensive models, differences in capability dominate.

\section{Avoiding These Patterns}
\label{sec:router}
We build a proof-of-concept router to show that the four patterns are avoidable. We choose \{Gemini 3.7 Flash, Opus 5 (high)\} as our roster, the same model pair we used in our random baseline. To predict difficulty, we compare five methods on Qwen3.5-27B by within-source $\gamma(\hat{d},d)$ and select SharedTrunkNet \citep{varshney2026llmrouterrethinkingrouting} (\S\ref{app:router}). A threshold on predicted difficulty determines the price point, with queries above the threshold routed to Opus 5 (high) and everything else to Gemini 3.7 Flash.

On the easy band, both models in our roster solve nearly all queries, and all routed accuracies are within noise of random routing (Figure~\ref{fig:g37-opus-pair}a). On the hard band, Opus 5 (high) leads Gemini 3.7 Flash by $7.2 \pm 7.6$~pp, which is within noise, so there is little room for accuracy gains through routing on this roster. Our router escalates hard-band queries more often than random ID, but the gap is mostly within noise OOD, suggesting that source exposure helps a router know when to escalate. Under the standard methodology, its accuracy is within noise of random routing at every budget (Table~\ref{tab:router-scores}).

When we replace Gemini 3.7 Flash with the much weaker gpt-oss-20b (low), both our router and the difficulty oracle underperform random routing at every budget, and the oracle does worst (Table~\ref{tab:ours-id}), since the hardest queries tend to also be the longest. Under our revised methodology, difficulty-based routing on this roster significantly outperforms random routing at most budgets on the easy band (Figure~\ref{fig:oss-opus-pair}).

Our router avoids the four patterns (Table~\ref{tab:router-patterns}). Its source AMI is comparable to that of the commercial routers, but its within-source $\gamma(d,a)$ is far higher, which indicates that it distinguishes queries within a source by difficulty. With a well-chosen roster, the primary advantage of routing is better-than-random escalation in distribution in the hard band, and the roster, not the router, determines nearly all of the accuracy.

\section{Discussion}
\label{sec:discussion}
In this work, we identify four patterns prevalent across commercial LLM routers: difficulty blindness, length reversal, semantic matching, and roster suboptimality. As a proof of concept, we develop a simple router between two models which avoids them, although it offers little measurable performance gain. Even with perfect knowledge of query difficulty, rosters larger than two models gain at most 1~pp, and on a well-chosen pair, a router that avoids the other three patterns still performs within noise of random routing. When a commercial router underperforms random routing, its roster is therefore the more likely cause.

Our analysis suggests that the problem of routing is a challenging one, without a reliable way of determining which queries to send to each model. Moreover, this challenge may become greater with future generations of models which are increasingly capable and optimized for long-horizon tasks. In that case, it may be even more difficult to predict based on the query alone whether the future cheap but (by today's standards) extremely capable models will fail a task.

\section*{Limitations}
\label{sec:limitations}
\paragraph{Single-turn routing.} We focus solely on single-turn routing, but this myopic approach applied per-turn can hurt performance since with KV caching, switching models is not free. We do not propose any solution to multi-turn routing; this paper focuses on single-turn routing and shows that, even in this simpler case, router designs are often suboptimal.

\paragraph{Dataset composition.} We attempted to make our training and evaluation datasets as diverse as possible, but we relied entirely on benchmarks because of their verifiability, and benchmark tasks may not resemble real user traffic. We only generate each outcome once, which can be noisy since model outcomes on a query vary between samples, and we did not test the stability of pass/fail outcomes over samples. Finally, our datasets only include English queries, although model capabilities can differ between languages \citep{shi2023language,ahuja-etal-2023-mega,singh-etal-2025-global,kang-etal-2026-multilingual}. We leave these explorations to future work.

\paragraph{Future models may specialize.} We only observe weak specialization among the models we tested. However, the model development landscape may shift to focus on specialized or personalized models whose capabilities vary greatly by domain, in which case future routers might benefit from larger rosters.

\paragraph{Weak difficulty predictions.} We find it difficult to predict query difficulty, especially OOD, which limits the efficacy of routing. Although we do not claim to have found the optimal difficulty predictor, its OOD performance within a source does not improve with a tenfold increase in training data (Figure~\ref{fig:data-scaling}), which suggests the limit lies in how much a query reveals about its difficulty rather than in our method or data. In addition, this task may become more challenging as model capabilities improve and it becomes increasingly difficult to predict which challenging tasks are harder than others.


\bibliography{custom}

\clearpage
\appendix

\section{Setup}
\label{app:setup}
\subsection{Dataset}
\begin{table*}[h!]
\small
\setlength{\tabcolsep}{4pt}
\caption{Data sources in our training and evaluation datasets, with the number of training and evaluation samples from each source.}
\label{tab:sources}
\begin{tabularx}{\textwidth}{l
    >{\hsize=0.9\hsize\raggedright\arraybackslash}X
    >{\hsize=1.1\hsize\raggedright\arraybackslash}X
    rr}
\toprule
& & & \multicolumn{2}{c}{Samples} \\
\cmidrule(lr){4-5}
Category & Source & Description & Train & Eval \\
\midrule
Coding      & HumanEval \citep{chen2021}                      & Python function synthesis with unit tests         & $82$  & $25$ \\
            & HumanEvalFix \citep{muennighoff2024octopack}    & Bug-fix coding tasks with unit tests              & $82$  & $25$ \\
            & MBPP \citep{austin2021}                         & Short Python programming tasks                    & $150$ & $25$ \\
            & BigCodeBench \citep{zhou2025}                   & Harder practical coding scenarios                 & $150$ & $25$ \\
\midrule
Instruction & IFEval \citep{zhou2023}                         & Instruction-following and formatting constraints  & $150$ & $100$ \\
\midrule
Knowledge   & MMLU \citep{hendrycks2021measuring}             & General knowledge MCQ                             & $150$ & $34$ \\
            & MMLU-Pro \citep{wang2024}                       & Harder MMLU-style benchmark                       & $150$ & $33$ \\
            & GPQA \citep{rein2024gpqa}                       & Graduate-level domain QA                          & $99$  & $33$ \\
\midrule
Math        & MATH \citep{math,lightman2024lets} & Competition-style math reasoning             & $150$ & $50$ \\
            & GSM8K \citep{cobbe2021}                         & Grade-school word problems                        & $150$ & $50$ \\
\midrule
QA          & Natural Questions \citep{kwiatkowski2019}       & Open-domain factual QA                            & $150$ & $50$ \\
            & TriviaQA \citep{joshi2017}                      & Trivia-style short-answer QA                      & $149$ & $50$ \\
\midrule
Office      & DataBench \citep{oses2024}                      & Office/data-analysis tasks                        & $150$ & $30$ \\
            & SpreadsheetBench \citep{ma2024}                 & Spreadsheet manipulation/reasoning                & $160$ & $40$ \\
            & GDPval \citep{patwardhan2026gdpval}             & Knowledge-work tasks                              & $120$ & $30$ \\
\midrule
Tool use    & BFCL \citep{patil2025}                          & Function/API tool-use                             & $250$ & $50$ \\
            & Tau-bench \citep{yao2025taubench}               & Interactive tool-use tasks                        & $100$ & $50$ \\
\midrule
HLE & Humanity's Last Exam \citep{phan2025lastexam} & Difficult exam-style reasoning (tools allowed) & $116$ & $100$ \\
\midrule
Total       &                                                 &                                                   & $2{,}508$ & $800$ \\
\bottomrule
\end{tabularx}
\end{table*}

Table~\ref{tab:sources} lists the sources for our dataset, grouped into eight diverse categories. We generate each model's response using the provider-default temperature. All benchmarks have verifiable rewards except for GDPval, which comprises just 3.8\% of the evaluation set, and Humanity's Last Exam, where the judge only has to compare the final answer with the answer key. For GDPval, we use the provided rubric, counting any response with a score of at least 0.5 as a pass. For Humanity's Last Exam, each model answers in an agentic loop with web search using the Tavily API and a sandboxed Python interpreter, for up to 15 tool calls. We block any search results from \url{huggingface.co}, \url{lastexam.ai}, and \url{github.com}, which may contain answers. Both of these datasets are judged using Opus 4.8.

\subsection{Model set}
Our first model set, selected based on the frontier in June 2026, comprises:
\begin{multicols}{2}
\small
    \begin{enumerate}
        \item Claude Haiku 4.5
        \item Claude Sonnet 4.6
        \item Claude Opus 4.8
        \item GPT 5.4
        \item GPT 5.4 Mini
        \item GPT 5.5
        \item GPT 5.5 (medium)
        \item gpt-oss-20b (low)
        \item gpt-oss-120b
        \item DeepSeek V3
        \item DeepSeek V4 Flash
        \item DeepSeek V4 Flash (thinking)
        \item DeepSeek V4 Pro
        \item DeepSeek V4 Pro (thinking)
        \item GLM-5.1
        \item GLM-5.2
        \item GLM-5.2 (high)
        \item Gemini 3.1 Flash
        \item Gemini 3.1 Flash (low)
        \item Gemini 3.1 Pro
    \end{enumerate}
\end{multicols}

Our second model set, revised in August 2026, incorporates six new model releases:
\begin{multicols}{2}
\small
    \begin{enumerate}
        \item DeepSeek V4 Flash 07-31 (thinking)
        \item GPT 5.6 Luna (high)
        \item Gemini 3.7 Flash
        \item Grok 4.6 (high)
        \item GPT 5.6 Sol (xhigh)
        \item Claude Opus 5 (high)
    \end{enumerate}
\end{multicols}
Note that we omitted Claude Fable 5 because it is substantially more expensive than any of the models in our list.

\subsection{Commercial routers}
We evaluate the following commercial routers:
\begin{enumerate}
    \item \textbf{OpenRouter:} Classifies the incoming query and routes based on which models are called most frequently for such tasks by OpenRouter users in the past seven days. We consider the medium, high, and xhigh cost bands. We evaluate both \texttt{openrouter/auto} and \texttt{openrouter/auto-beta}, where the latter is an early-access track with ``new routing behaviors.''
    \item \textbf{Azure Model Router:} Analyzes the query and selects a model by ``[estimating] which model in the pool delivers the best result for [the] specific prompt.'' We assess the balanced and quality settings.
    \item \textbf{Not Diamond:} Similar to Azure Model Router, choosing a model based on cost and estimated performance. It offers cost and quality settings, and we evaluate both.
    \item \textbf{vLLM Semantic Router:} Uses a fine-tuned ModernBERT model to assign each query to a user-defined category, and selects the model with the highest observed accuracy on that category. We define categories using the eight categories (not data sources) in our data, and we feed in all our training data from our 26-model set.
    \item \textbf{Nadir Router:} Routes based on cost and estimated performance using customizable model configurations for three tiers: simple, medium, and complex. The medium tier is optional, so the router functions as a two- or three-tier escalation ladder, making it the closest analog to our routing policies for comparison. We assess two versions: two-tier with Claude Haiku 4.5 and Opus 5, and three-tier with Claude Haiku 4.5, Sonnet 5, and Opus 5.
    \item \textbf{OrcaRouter:} A multi-armed bandit router described in \citet{bao2026}. Its adaptive mode uses an online LinUCB algorithm to ``[converge] on the best option from real production traffic, weighing quality, cost, latency, and reliability.'' The adaptive mode requires a warm-up phase, so we initially send 1{,}000 training samples before the evaluation samples so the router has time to converge first.
\end{enumerate}
Among these, the OpenRouter routers, vLLM Semantic Router, and Nadir are not perfect comparisons with our router. The OpenRouter router families choose the most popular model for each task type, meaning they route by popularity instead of predicted quality. The vLLM Semantic Router only focuses on accuracy and offers no cost-saving mechanism. Nadir does not choose its own roster. At the time of access, Nadir only offered Haiku 4.5, Sonnet 5, Opus 5, and Fable 5 as available models for each tier, and the user must configure the ladder. We used \{Haiku 4.5, Opus 5\} and \{Haiku 4.5, Sonnet 5, Opus 5\} because these are the configurations whose costs fall in the same range as the other commercial routers. \{Sonnet 5, Opus 5\} would be more expensive, and Fable 5 more expensive than Opus 5.

\paragraph{Router calling.} We accessed each router between August 25--26, 2026. Every router, including our own, received the same input: the query text as a single user message without tool schemas, since the routing endpoints reject or ignore them. For each commercial router's output, we extract the chosen model and generate our own completion using OpenRouter, grading our generated output and storing the cost. This process ensures consistency across all evaluated routers. For models outside our model set, we estimate the per-model mean cost $\overline{c}_i$ from 100 queries stratified by source rather than from the full evaluation set.

The Azure content filter rejected three queries across IFEval, MMLU, and MMLU-Pro, so we compute its scores and our baseline comparison across the remaining 797 queries. Orca routed 37 queries to models not available on OpenRouter. For seven of these (5 for gpt-5-search-api and 2 for gpt-3.5-turbo-1106), we successfully called them using the OpenAI API. The remaining 30 (gpt-5-chat-latest, gpt-5.1-chat-latest, gpt-5.2-chat-latest, gemini-robotics-er-1.6-preview, qwen3-max-preview) are deprecated across both platforms, and we score the router on the remaining 770. Given that these models are deprecated, it is unlikely that they would have improved Orca's routing performance.

\paragraph{Omitted routers.} The Cursor, Factory, and Entelligence routers are all designed to select models per-turn within multi-turn coding agent sessions, whereas our routers make per-query decisions over general traffic. The Ramp Router does not route per query. Merge Gateway focuses on live cost, throughput, and latency conditions, rather than per-query predicted success rates. Sakana Fugu \citep{tang2026sakanafugutechnicalreport} predicts answer success but does not expose the routing decision, making it difficult to accurately compare its underlying model costs with our router. Its Sakana Fugu Ultra counterpart is an orchestrator that handles more than just routing. AWS Bedrock Intelligent Prompt Routing uses a customizable two-tier scheme, but the only available models are outdated (Claude 3.5, Llama 3.x, and Amazon Nova series).

\subsection{Router training}
\paragraph{ID and OOD fits.} We train the difficulty predictor in two ways. In the in-distribution (ID) setting, we fit one predictor using the full training set, which covers the same eight categories as the evaluation set. In the out-of-distribution (OOD) setting, we fit eight predictors, each on the training data from seven categories, and score each evaluation query with the predictor that never saw that category. The OOD scores are therefore the concatenation of eight separately fit models, and we apply a single threshold across all of them, making this setting doubly conservative: not only are we evaluating OOD, but also the eight fits are not calibrated relative to each other, so a threshold tuned for one can over- or under-escalate on the other. The threshold is chosen so the router's cost on the evaluation set equals the stated price point, which depends on whether we use realized or per-model mean costs. Note that this uses evaluation costs but not evaluation outcomes.

\section{Definitions}
\label{app:def}

\paragraph{Goodman-Kruskal's gamma.} For two ordinal variables, consider every pair of queries that differs on both. A pair is \emph{concordant} if the two variables order the pair the same way, and \emph{discordant} otherwise. $\gamma$ is the number of concordant minus discordant pairs divided by their sum. Pairs tied on either variable are excluded. By comparison, Spearman's $\rho$ treats tied values as sharing a rank, so when the dependent variable takes on fewer discrete values (as in the case of our router, with only two models in the roster), the maximum achievable value is $\sqrt{3r(1-r)}$, where $r$ is the escalation rate. At 10\% escalation rate, a perfect two-model router achieves only $\rho \approx 0.52$ but $\gamma=1$.

\paragraph{Source AMI.} Source adjusted mutual information (AMI) measures how much of the uncertainty in the router's choice is explained by the query's source. We arrange the router's choices as a table of counts, with one row per source and one column per chosen model. Let $n_{si}$ be the number of queries from source $s$ sent to model $i$, $n_s = \sum_i n_{si}$, $n_i = \sum_s n_{si}$, and $n = \sum_{s,i} n_{si}$. The entropy $H$ of the chosen models and the mutual information $I$ between source and chosen model are
\begin{align*}
  H &= -\sum_{i} \frac{n_i}{n} \log \frac{n_i}{n}, \\
  I &= \sum_{s} \sum_{i} \frac{n_{si}}{n} \log \frac{n\, n_{si}}{n_s\, n_i},
\end{align*}
where terms with $n_{si} = 0$ are zero. Since $0 \le I \le H$, the AMI is given by
$$\frac{I-\mathbb{E}(I)}{H-\mathbb{E}(I)},$$
where $\mathbb{E}(I)$ is the expected mutual information if the queries were shuffled at random between models while keeping every source's size and every model's count fixed. This way, the AMI is 0 in expectation if knowing the source tells us nothing about the router's picks, and 1 if the source fully determines the model.

\section{Router Design}
\label{app:router}
We explore a range of methods for predicting query difficulty, using Qwen3.5-27B as the base model since the Qwen family of models is known to be competitive at predicting difficulty \citep{cencerrado2026answerneededpredictingllm,lee2026probingdifficultyperceptionmechanism,varshney2026llmrouterrethinkingrouting}. We choose the 27B variant because a full 27B prefill is more costly than a smaller probe, but still a fraction of the inference cost of any model in the roster. We encode each query using Qwen's chat template. We analyze a range of methods using final-token activations to estimate query difficulty:
\begin{enumerate}
    \item \textbf{Linear probe:} Train a single linear head on a layer's activations to predict the failure rate \citep{lee2026probingdifficultyperceptionmechanism};
    \item \textbf{Difference-of-means probe:} Compute the mean-correct minus mean-incorrect activation direction for each model, and average the projections \citep{cencerrado2026answerneededpredictingllm};
    \item \textbf{Linear RankNet:} Fit a linear scoring direction using source-balanced pairs of questions whose failure rates differ by at least 0.3 and optimize the pairwise logistic ranking loss \citep{10.1145/1102351.1102363};
    \item \textbf{PCA logistic probes:} For each model, select an upper-half layer using Fisher separation, reduce it to 100 principal components, and fit a correctness classifier, outputting the average over all model pass probabilities \citep{varshney2026llmrouterrethinkingrouting};
    \item \textbf{SharedTrunkNet:} Concatenate all model-specific PCA-100 representations and train a neural network across ten seeds, averaging the best five ensembles \citep{varshney2026llmrouterrethinkingrouting}.
\end{enumerate}

\begin{table*}[t]
    \small
    \centering
    \setlength{\tabcolsep}{4pt}
    \caption{Comparison between difficulty prediction methods by global and mean per-source $\gamma(\hat{d},d)$, the Goodman-Kruskal correlation between predicted and true query difficulty. We consider both the in-distribution (ID) and out-of-distribution (OOD) fits, described in Appendix~\ref{app:setup}. Confidence intervals are 95\% jackknife intervals over evaluation queries. Bold marks the highest point estimate in each column.}
    \label{tab:difficulty-comp}
    \begin{tabular}{lcccc}
    \toprule
    & \multicolumn{2}{c}{ID} & \multicolumn{2}{c}{OOD} \\
    \cmidrule(lr){2-3} \cmidrule(lr){4-5}
    Method & Global & Source & Global & Source \\
    \midrule
    Linear probe        & \ci[0.00]{0.29}{0.05}  & \ci[0.00]{0.09}{0.07}  & \ci[-0.00]{0.09}{0.05}  & \ci[0.00]{0.05}{0.07} \\
    Difference of means & \ci[0.00]{0.41}{0.04}  & \ci[0.00]{0.27}{0.07}  & \ci[-0.00]{-0.03}{0.05} & \ci[0.00]{0.25}{0.07} \\
    RankNet             & \ci[0.00]{0.28}{0.05}  & \ci[0.00]{0.28}{0.07}  & \cib[-0.00]{0.23}{0.05} & \ci[0.00]{0.09}{0.08} \\
    PCA logistic        & \ci[0.00]{0.58}{0.04}  & \ci[0.00]{0.38}{0.07}  & \ci[-0.00]{0.14}{0.05}  & \cib[0.00]{0.29}{0.07} \\
    SharedTrunkNet      & \cib[0.00]{0.59}{0.04} & \cib[0.00]{0.39}{0.07} & \ci[-0.00]{0.12}{0.05}  & \ci[0.00]{0.28}{0.07} \\
    \bottomrule
    \end{tabular}
\end{table*}

PCA logistic and SharedTrunkNet are tied within noise on each metric. Since we cannot distinguish between these two methods on our data, we defer to the reasoning in \citet{varshney2026llmrouterrethinkingrouting}, which argues for SharedTrunkNet because it shares information across targets, calibrates scores on a comparable scale, and computes correctness probabilities in a single forward pass.

Interestingly, the highest achievable global $\gamma(\hat{d},d)$ is substantially lower OOD than ID, whereas the per-source correlation suffers less. This observation suggests that the primary benefit of seeing a source in training is learning its average difficulty, since the probe's ability to order queries within a source generalizes relatively well.

\section{Supplementary Tables}
\label{app:tables}
This section contains supplementary tables to the main text.
\begin{table*}
    \small
    \centering
    \caption{Costs and accuracies of commercial routers on our evaluation dataset. \emph{Chosen} is the number of distinct models the router chose, and \emph{Roster} is the number of models in its roster. OR: OpenRouter; ND: Not Diamond; SR: Semantic Router.}
    \label{tab:commercial-comp}
    \begin{threeparttable}
    \begin{tabular}{l S[table-format=3.2] S[table-format=2.1] rr}
    \toprule
    & & & \multicolumn{2}{c}{Models} \\
    \cmidrule(lr){4-5}
    Router & {Cost (\$/1k q.)} & {Acc.\ (\%)} & Chosen & Roster \\
    \midrule
    OR auto (med)           &  42.95 & 76.1 & $7$  & $415$\rlap{\tnote{$*$}} \\
    OR auto (high)          &  68.83 & 77.9 & $10$ & $415$\rlap{\tnote{$*$}} \\
    OR auto (xhigh)         &  59.04 & 78.5 & $7$  & $415$\rlap{\tnote{$*$}} \\
    OR auto-$\beta$ (med)   &   9.54 & 73.5 & $7$  & $415$\rlap{\tnote{$*$}} \\
    OR auto-$\beta$ (high)  &  23.50 & 72.5 & $6$  & $415$\rlap{\tnote{$*$}} \\
    OR auto-$\beta$ (xhigh) &  56.62 & 79.8 & $10$ & $415$\rlap{\tnote{$*$}} \\
    Azure (balanced)        &  15.69 & 72.8 & $13$ & $21$ \\
    Azure (quality)         &  52.09 & 79.3 & $10$ & $21$ \\
    ND (cost)               &  22.91 & 72.5 & $5$  & $132$ \\
    ND (quality)            &  27.41 & 77.2 & $5$  & $132$ \\
    vLLM SR                 &  68.97 & 80.9 & $6$  & $26$ \\
    Nadir (2-tier)          & 119.87 & 81.1 & $2$  & $2$\rlap{\tnote{$\dagger$}} \\
    Nadir (3-tier)          & 114.83 & 80.6 & $3$  & $3$\rlap{\tnote{$\dagger$}} \\
    Orca (adaptive)         &  29.23 & 71.0 & $96$ & $187$ \\
    \bottomrule
    \end{tabular}
    \begin{tablenotes}
        \small
        \item[$*$] The roster size for OpenRouter routers is unknown, and 415 reflects the full pool of models available on OpenRouter at the time of evaluation.
        \item[$\dagger$] Nadir's roster size equals the number of tiers we configured, not the full list of models the platform supports.
    \end{tablenotes}
    \end{threeparttable}
\end{table*}

\begin{table*}
    \small
    \centering
    \setlength{\tabcolsep}{4pt}
    \caption{Our router evaluated as in Table~\ref{tab:router-patterns}, using accuracy over the entire evaluation set with realized generation costs. \emph{ID} and \emph{OOD} are the in-distribution and out-of-distribution fits. \emph{Difficulty oracle} is a router with perfect knowledge of query difficulty that escalates the hardest queries first. Intervals are 95\% paired Wald intervals. A bold point estimate indicates that the router is not significantly below the random router.}
    \label{tab:router-scores}
    \begin{tabular}{l S[table-format=2.1] ccc}
    \toprule
    & & \multicolumn{3}{c}{$\Delta$ vs.\ random router (pp)} \\
    \cmidrule(lr){3-5}
    Price point & {Random acc.\ (\%)} & Ours (ID) & Ours (OOD) & Difficulty oracle \\
    \midrule
    \multicolumn{5}{@{}l}{\textit{Gemini 3.7 Flash $\rightarrow$ Claude Opus 5 (high)}} \\
    \addlinespace
    \$10/1k q. & 81.2 & \cib[-0.0]{-0.1}{0.3} & \cib[-0.0]{-0.2}{0.2} & \cib[-0.0]{-0.1}{0.1} \\
    \$20/1k q. & 81.3 & \cib[-0.0]{-0.2}{0.4} & \cib[-0.0]{-0.7}{0.7} & \cib[-0.0]{-0.2}{0.2} \\
    \$40/1k q. & 81.5 & \cib[-0.0]{-0.4}{0.8} & \cib[-0.0]{-0.7}{1.0} & \cib[-0.0]{+0.2}{0.9} \\
    \midrule
    \multicolumn{5}{@{}l}{\textit{gpt-oss-20b (low) $\rightarrow$ Claude Opus 5 (high)}} \\
    \addlinespace
    \$10/1k q. & 57.4 & \ci[-0.0]{-1.5}{0.5} & \cib[-0.0]{-0.7}{0.7} & \ci[-0.0]{-1.9}{0.2} \\
    \$20/1k q. & 59.3 & \ci[-0.0]{-2.8}{0.8} & \ci[-0.0]{-1.9}{1.0} & \ci[-0.0]{-3.8}{0.5} \\
    \$40/1k q. & 63.2 & \ci[-0.0]{-4.6}{1.3} & \cib[-0.0]{-0.7}{1.7} & \ci[-0.0]{-6.4}{1.1} \\
    \bottomrule
    \end{tabular}
\end{table*}

\begin{table*}
    \small
    \centering
    \setlength{\tabcolsep}{4pt}
    \caption{Comparison between commercial routers and random routers at matched costs, as in Table~\ref{tab:router-patterns}, but also matched to models available at the time of each commercial router's last update. Every router's roster includes the GPT 5.6x series or models released after, and some include Gemini 3.7 Flash, which was released later. Commercial routers whose rosters do not contain Gemini 3.7 Flash at the time of access are compared against \{GPT 5.6 Luna (high), GPT 5.6 Sol (xhigh)\}, while those that contain it are compared against \{Gemini 3.7 Flash, Opus 5 (high)\}. The shaded cell in each row is the comparison matched to that router's roster. A bold point estimate indicates that the router is not significantly below the random router. Five routers are within noise of a random router using models available at the time, and all other routers underperform.}
    \label{tab:commercial-full}
    \begin{tabular}{l S[table-format=3.2] S[table-format=2.1] cc}
    \toprule
    & & & \multicolumn{2}{c}{$\Delta$ vs.\ random router (pp)} \\
    \cmidrule(lr){4-5}
    Router & {Cost (\$/1k q.)} & {Acc.\ (\%)} & Luna / Sol & Flash / Opus \\
    \midrule
    OR auto (med)           &  42.95 & 76.1 & \ci[-0.0]{-3.9}{2.2}  & \cellcolor{gray!15}\ci[-00.0]{-5.5}{2.1} \\
    OR auto (high)          &  68.83 & 77.9 & \ci[-0.0]{-2.9}{2.3}  & \cellcolor{gray!15}\ci[-00.0]{-4.0}{2.0} \\
    OR auto (xhigh)         &  59.04 & 78.5 & \cib[-0.0]{-2.2}{2.2} & \cellcolor{gray!15}\ci[-00.0]{-3.3}{2.0} \\
    OR auto-$\beta$ (med)   &   9.54 & 73.5 & \ci[-0.0]{-4.1}{2.5}  & \cellcolor{gray!15}\ci[-00.0]{-7.7}{2.5} \\
    OR auto-$\beta$ (high)  &  23.50 & 72.5 & \ci[-0.0]{-6.1}{2.3}  & \cellcolor{gray!15}\ci[-00.0]{-8.8}{2.5} \\
    OR auto-$\beta$ (xhigh) &  56.62 & 79.8 & \cib[-0.0]{-1.0}{1.8} & \cellcolor{gray!15}\ci[-00.0]{-2.0}{1.9} \\
    Azure (balanced)        &  15.69 & 72.8 & \cellcolor{gray!15}\ci[-0.0]{-5.1}{2.5}  & \ci[-00.0]{-8.4}{2.6} \\
    Azure (quality)         &  52.09 & 79.3 & \cellcolor{gray!15}\cib[-0.0]{-1.3}{1.7} & \ci[-00.0]{-2.2}{1.9} \\
    ND (cost)               &  22.91 & 72.5 & \cellcolor{gray!15}\ci[-0.0]{-6.0}{2.5}  & \ci[-00.0]{-8.8}{2.7} \\
    ND (quality)            &  27.41 & 77.2 & \cellcolor{gray!15}\cib[-0.0]{-1.6}{2.1} & \ci[-00.0]{-4.1}{2.2} \\
    vLLM SR                 &  68.97 & 80.9 & \cib[-0.0]{+0.1}{2.1} & \cellcolor{gray!15}\cib[-00.0]{-1.0}{1.8} \\
    Nadir (2-tier)          & 119.87 & 81.1 & \cellcolor{gray!15}\cib[-0.0]{+0.4}{2.4} & \ci[-00.0]{-1.4}{1.1} \\
    Nadir (3-tier)          & 114.83 & 80.6 & \cellcolor{gray!15}\cib[-0.0]{-0.1}{2.4} & \ci[-00.0]{-1.8}{1.2} \\
    Orca (adaptive)         &  29.23 & 71.0 & \cellcolor{gray!15}\ci[-0.0]{-8.0}{2.6}  & \ci[-00.0]{-10.5}{2.8} \\
    \bottomrule
    \end{tabular}
\end{table*}

\begin{table*}
    \centering
    \small
    \setlength{\tabcolsep}{4pt}
    \caption{Pattern metrics for our router fit in distribution and for a difficulty oracle that escalates the hardest queries by true difficulty, at the same three price points as Table~\ref{tab:router-patterns}. The oracle's $\gamma(d,a)$ is 1 by construction. Intervals are 95\% jackknife intervals over evaluation queries. A bold point estimate indicates a ``pass,'' as in Table~\ref{tab:router-patterns}.}
    \label{tab:ours-id}
    \begin{tabular}{lcccc}
    \toprule
    & \makecell{Difficulty\\blindness}
    & \makecell{Length\\reversal}
    & \multicolumn{2}{c}{Semantic matching} \\
    \cmidrule(lr){2-2} \cmidrule(lr){3-3} \cmidrule(lr){4-5}
    Router
    & $\gamma(d,a)$
    & $\gamma(\ell,c)$
    & Source AMI
    & \makecell{Within-source\\$\gamma(d,a)$} \\
    \midrule
    Ours, ID (\$10/1k q.)          & \cib{+0.81}{0.08} & \cib{+0.92}{0.04} & \ci{0.46}{0.07} & \ci{+0.24}{0.29} \\
    Ours, ID (\$20/1k q.)          & \cib{+0.76}{0.06} & \cib{+0.67}{0.12} & \ci{0.49}{0.07} & \cib{+0.20}{0.19} \\
    Ours, ID (\$40/1k q.)          & \cib{+0.74}{0.05} & \cib{+0.61}{0.08} & \ci{0.56}{0.06} & \cib{+0.31}{0.18} \\
    \midrule
    Difficulty oracle (\$10/1k q.) & \cib{+1.00}{0.00} & \cib{+0.25}{0.27} & \ci{0.19}{0.11} & \cib{+1.00}{0.00} \\
    Difficulty oracle (\$20/1k q.) & \cib{+1.00}{0.00} & \cib{+0.31}{0.15} & \ci{0.25}{0.08} & \cib{+1.00}{0.00} \\
    Difficulty oracle (\$40/1k q.) & \cib{+1.00}{0.00} & \cib{+0.51}{0.09} & \ci{0.31}{0.07} & \cib{+1.00}{0.00} \\
    \bottomrule
    \end{tabular}
\end{table*}

\section{Supplementary Figures}
\label{app:figures}
This section contains supplementary figures to the main text.
\begin{figure}[!ht]
    \centering
    \includegraphics[width=0.8\linewidth]{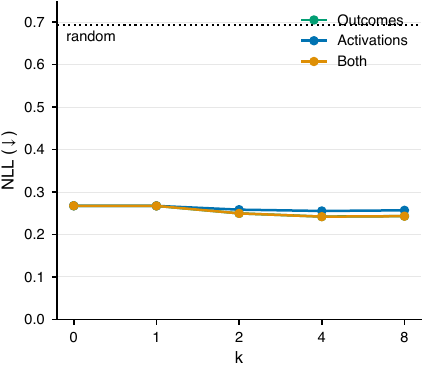}
    \caption{Negative log-likelihood (NLL) of the MIRT model as the number of interaction dimensions $k$ grows. The fits use observed pass rates (\emph{outcomes}), Qwen3.5-27B activations (\emph{activations}), or a combination of these (\emph{both}). $k=0$ is the Rasch model. 2 standard error bars do not exceed 0.02, and the lines for \emph{outcomes} and \emph{both} are identical.}
    \label{fig:mirt}
\end{figure}

\begin{figure}[h!]
    \centering
    \includegraphics[width=0.9\linewidth]{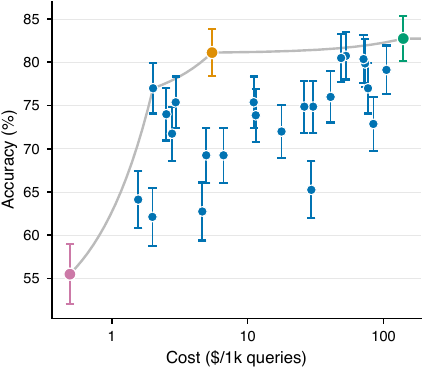}
    \caption{Standalone costs and accuracies of each model in the 26-model set. The gray line is the cost-accuracy frontier, which linearly interpolates between models on the Pareto frontier. gpt-oss-20b (low) is marked in pink, Gemini 3.7 Flash in yellow, and Opus 5 (high) in green. Error bars are 95\% confidence intervals over queries.}
    \label{fig:paretos}
\end{figure}

\begin{figure*}[!ht]
    \centering
    \includegraphics[width=0.6\linewidth]{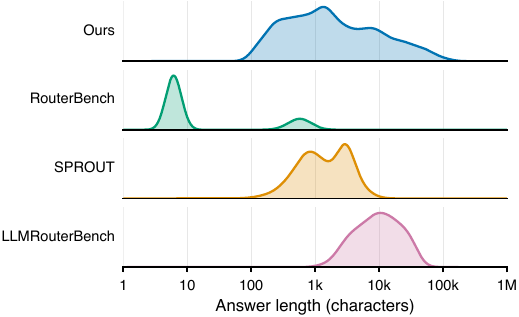}
    \caption{Histogram of average answer length $\ell_j$ in characters for each dataset. Most of RouterBench's queries are multiple-choice questions. Our dataset spans the widest range of answer lengths, which may be why no previous evaluation reported length reversal.}
    \label{fig:length-hist}
\end{figure*}

\begin{figure*}[!ht]
    \centering
    \begin{subfigure}{0.49\linewidth}
        \centering\includegraphics[width=0.8\linewidth]{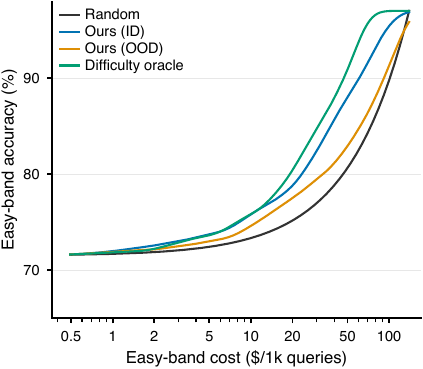}
        \caption{Easy-band accuracy against easy-band cost.}
    \end{subfigure}
    \begin{subfigure}{0.49\linewidth}
        \centering\includegraphics[width=0.8\linewidth]{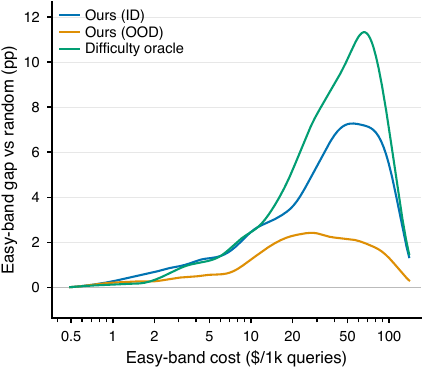}
        \caption{Easy-band accuracy gap with random routing.}
    \end{subfigure}
    \caption{(a) is a replication of Figure~\ref{fig:g37-opus-pair}(a) with \{gpt-oss-20b (low), Opus 5 (high)\} as the roster. (b) shows the improvement of difficulty-based routing methods against the random routing baseline on the easy band. Note that the hard-band escalation against easy-band cost figure is the same shape as in Figure~\ref{fig:g37-opus-pair}(b), since we are using the same router, just with a different roster. On (a), 95\% half-widths do not exceed 3.6~pp; on (b), 95\% half-widths do not exceed 2.5~pp, and both ours (ID) and difficulty oracle achieve statistically significant accuracy gaps when easy-band costs range from \$6--139/1k queries, while ours (OOD) achieves significant gains over most of \$11--64/1k queries.}
    \label{fig:oss-opus-pair}
\end{figure*}

\begin{figure*}[!ht]
    \centering
    \includegraphics[width=0.7\linewidth]{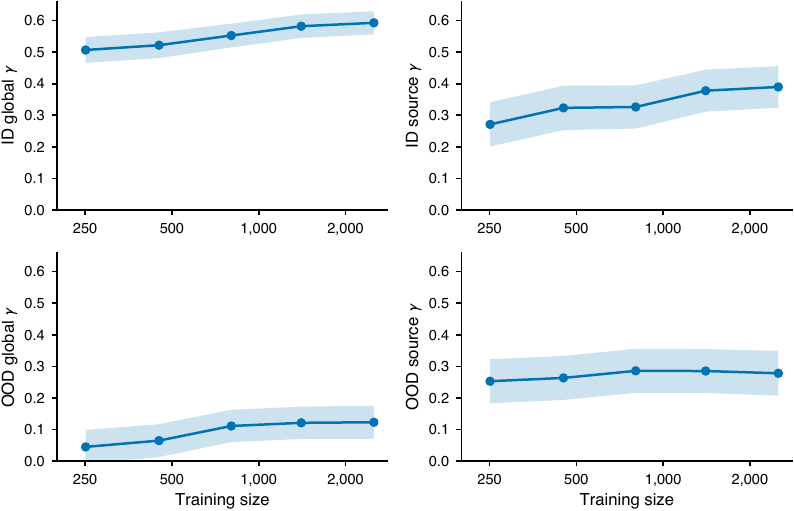}
    \caption{Scaling of $\gamma(\hat{d},d)$ predictions using the SharedTrunkNet architecture as a function of the training size. We plot both global and source correlations ID and OOD, with 95\% jackknife confidence intervals. For points trained on less than the full training set, we average over three seeds.}
    \label{fig:data-scaling}
\end{figure*}

\begin{figure*}
    \centering
    \includegraphics[width=0.9\linewidth]{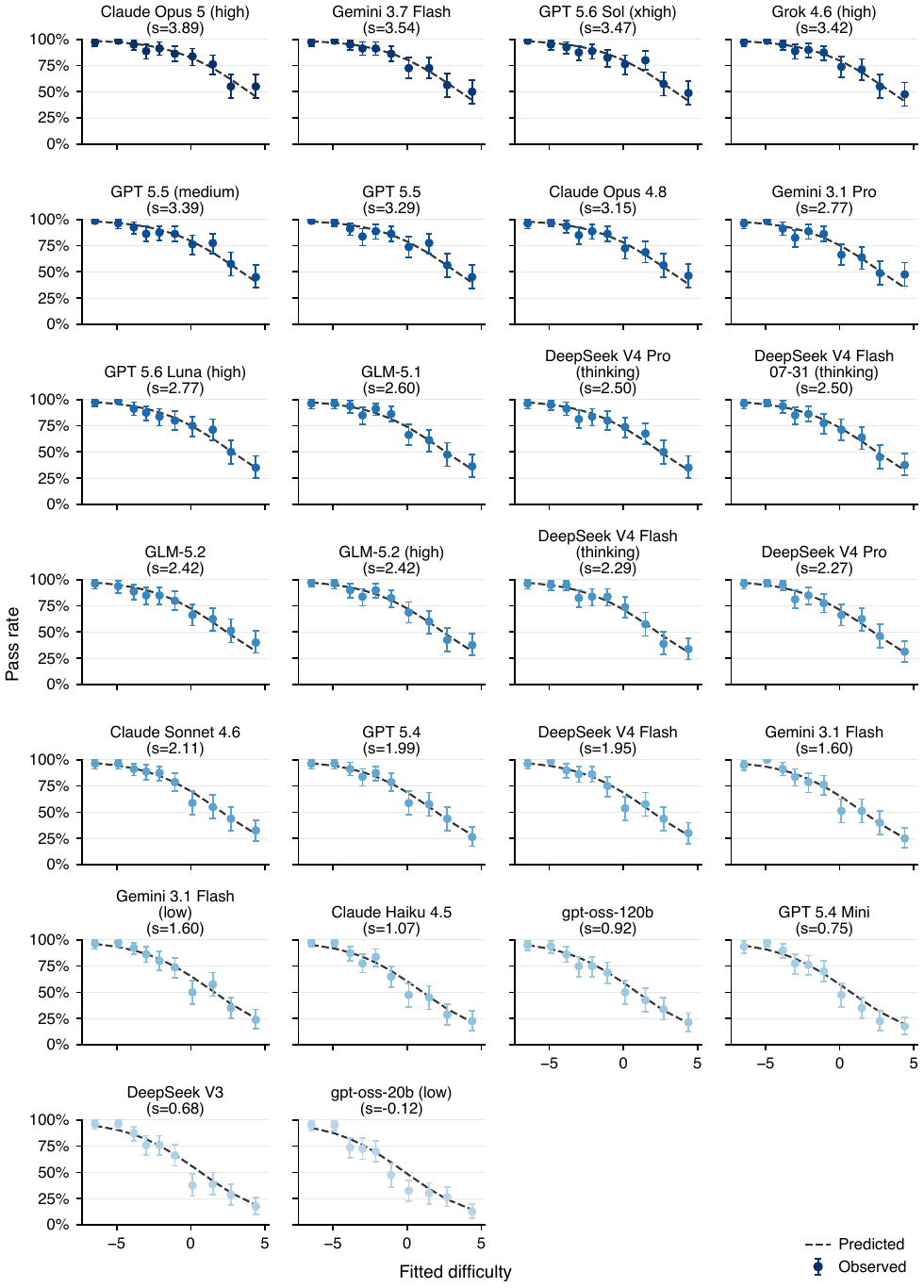}
    \caption{Observed versus predicted pass rates for each model using the Rasch model. We group evaluation queries into ten bins by fitted difficulty, so each point represents a difficulty band. Error bars are 95\% bootstrapped confidence intervals. Because all models are evaluated on the same queries with the same difficulty predictions, each predicted line is linearly offset in logit space. The fact that each predicted line closely matches observed outcomes indicates that the Rasch model fits outcomes well.}
    \label{fig:rasch}
\end{figure*}


\begin{figure*}
    \centering
    \begin{subfigure}{0.49\linewidth}
        \centering
        \includegraphics[width=0.72\linewidth]{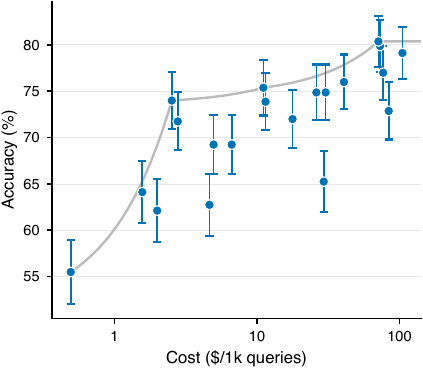}
        \caption{20-model set: standalone models.}
    \end{subfigure}
    \begin{subfigure}{0.49\linewidth}
        \centering
        \includegraphics[width=0.72\linewidth]{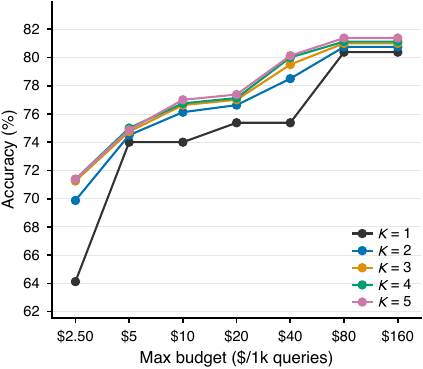}
        \caption{20-model set: router accuracies.}
    \end{subfigure}
    \begin{subfigure}{0.49\linewidth}
        \centering
        \includegraphics[width=0.72\linewidth]{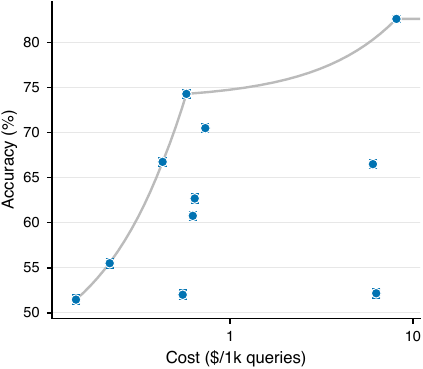}
        \caption{RouterBench: standalone models.}
    \end{subfigure}
    \begin{subfigure}{0.49\linewidth}
        \centering
        \includegraphics[width=0.72\linewidth]{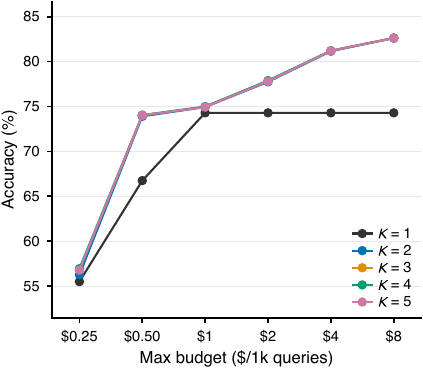}
        \caption{RouterBench: router accuracies.}
    \end{subfigure}
    \begin{subfigure}{0.49\linewidth}
        \centering
        \includegraphics[width=0.72\linewidth]{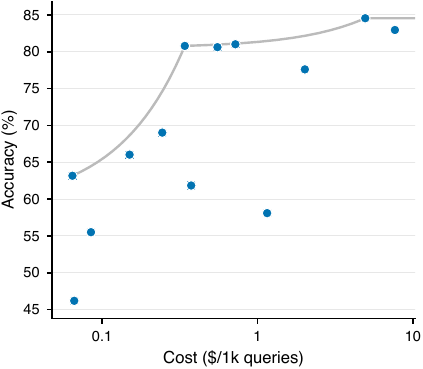}
        \caption{SPROUT: standalone models.}
    \end{subfigure}
    \begin{subfigure}{0.49\linewidth}
        \centering
        \includegraphics[width=0.72\linewidth]{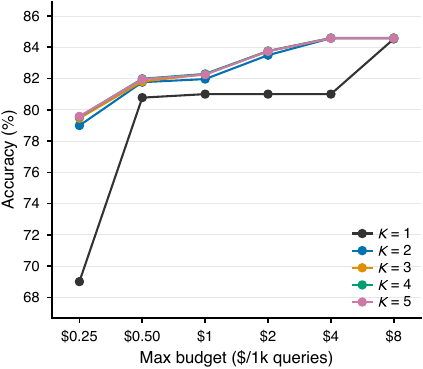}
        \caption{SPROUT: router accuracies.}
    \end{subfigure}
    \begin{subfigure}{0.49\linewidth}
        \centering
        \includegraphics[width=0.72\linewidth]{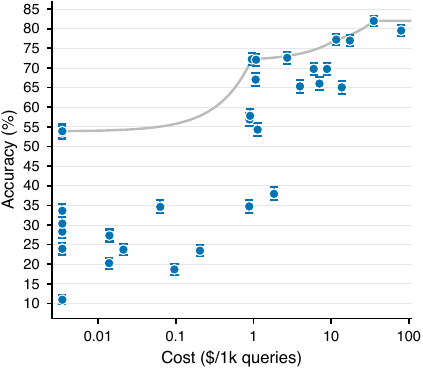}
        \caption{LLMRouterBench: standalone models.}
    \end{subfigure}
    \begin{subfigure}{0.49\linewidth}
        \centering
        \includegraphics[width=0.72\linewidth]{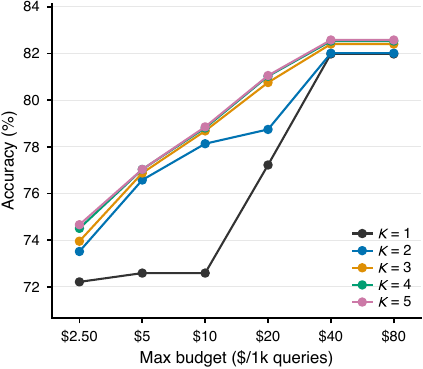}
        \caption{LLMRouterBench: router accuracies.}
    \end{subfigure}
    \caption{Replication of Figures~\ref{fig:paretos} and \ref{fig:roster-bloat} for the 20-model set, RouterBench, SPROUT, and LLMRouterBench, demonstrating that small roster sizes suffice. Note that some models in LLMRouterBench are listed with \$0.00 inference costs. Half-widths for roster accuracies do not exceed 3.5~pp (20-model set), 0.5~pp (RouterBench), 0.4~pp (SPROUT), and 1.5~pp (LLMRouterBench).}
    \label{fig:granularity-full}
\end{figure*}

\begin{figure*}
    \centering
    \begin{subfigure}{0.49\linewidth}
        \centering\includegraphics[width=0.75\linewidth]{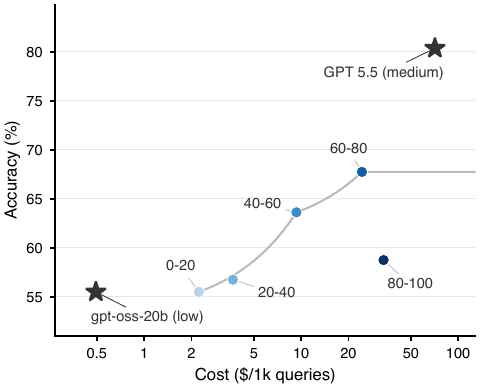}
        \caption{20-model set: difficulty-based escalation.}
    \end{subfigure}
    \begin{subfigure}{0.49\linewidth}
        \centering\includegraphics[width=0.75\linewidth]{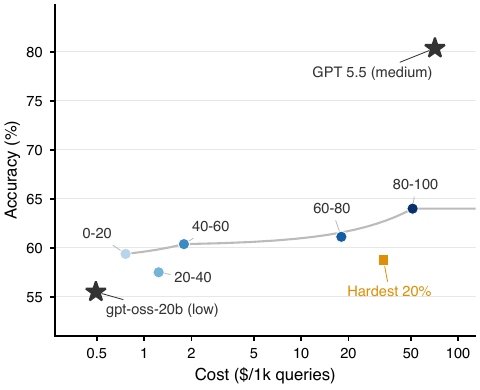}
        \caption{20-model set: length-based escalation.}
    \end{subfigure}
    \begin{subfigure}{0.49\linewidth}
        \centering\includegraphics[width=0.75\linewidth]{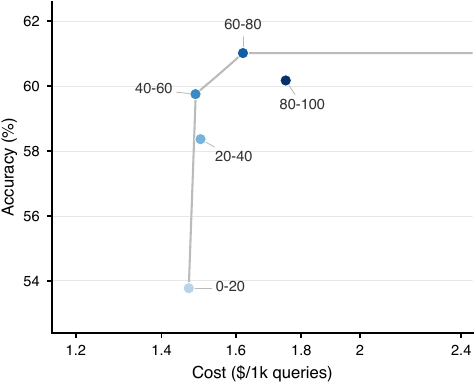}
        \caption{RouterBench: difficulty-based escalation.}
    \end{subfigure}
    \begin{subfigure}{0.49\linewidth}
        \centering\includegraphics[width=0.75\linewidth]{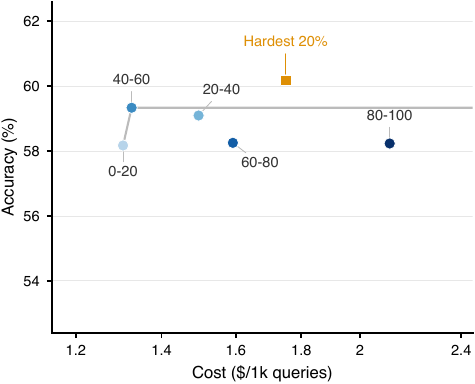}
        \caption{RouterBench: length-based escalation.}
    \end{subfigure}
    \begin{subfigure}{0.49\linewidth}
        \centering\includegraphics[width=0.75\linewidth]{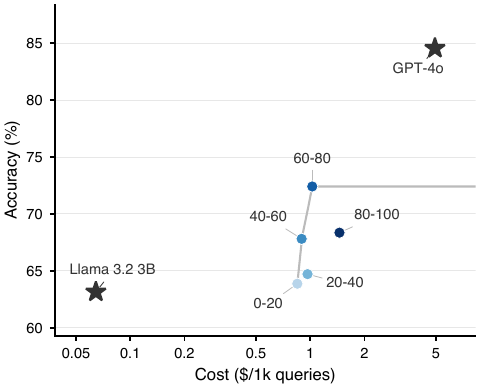}
        \caption{SPROUT: difficulty-based escalation.}
    \end{subfigure}
    \begin{subfigure}{0.49\linewidth}
        \centering\includegraphics[width=0.75\linewidth]{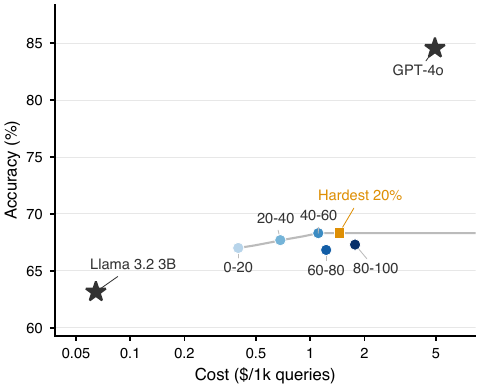}
        \caption{SPROUT: length-based escalation.}
    \end{subfigure}
    \begin{subfigure}{0.49\linewidth}
        \centering\includegraphics[width=0.75\linewidth]{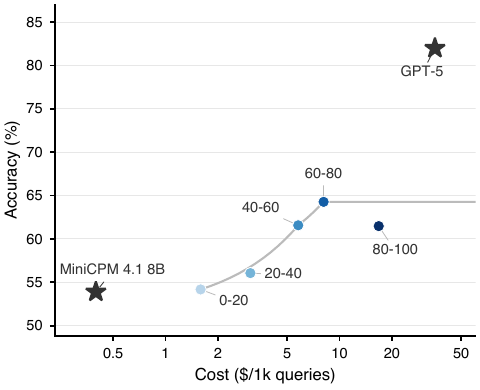}
        \caption{LLMRouterBench: difficulty-based escalation.}
    \end{subfigure}
    \begin{subfigure}{0.49\linewidth}
        \centering\includegraphics[width=0.75\linewidth]{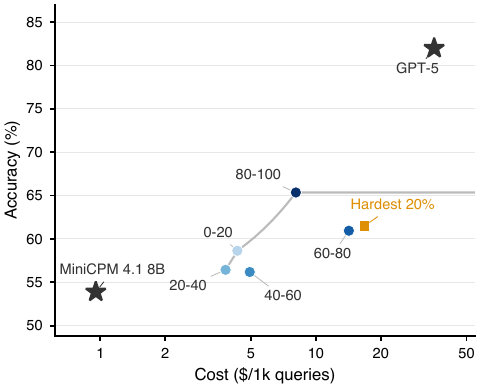}
        \caption{LLMRouterBench: length-based escalation.}
    \end{subfigure}
    \caption{Replication of Figure~\ref{fig:escalation} for all other datasets. We pair the cheapest model with the top-scoring model so accuracy and cost differences are most visible. For RouterBench, we route between Mistral 7B and GPT-4, and we do not include the endpoint stars because they are too far from the area of focus. The trends discussed in Section~\ref{sec:eval} hold across datasets except for RouterBench on length-based escalation, where query outputs differ little in length so there is less headroom for accuracy hacking (Figure~\ref{fig:length-hist}). The maximum half-widths for each dataset are 3.5~pp (20-model set), 0.9~pp (RouterBench), 0.4~pp (SPROUT), and 1.7~pp (LLMRouterBench). Escalating the 60--80 difficulty band significantly exceeds 80--100 on accuracy on every dataset except RouterBench.}
    \label{fig:escalation-full}
\end{figure*}

\begin{figure*}
    \centering
    \begin{subfigure}{0.49\linewidth}
        \centering
        \includegraphics[width=0.7\linewidth]{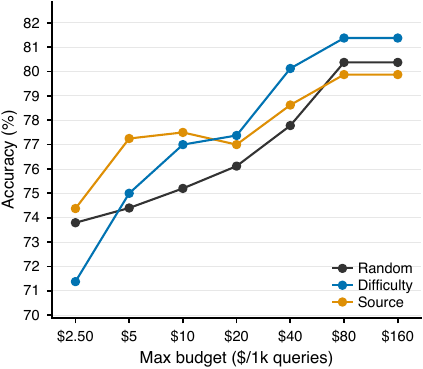}
        \caption{20-model set: realized costs.}
    \end{subfigure}
    \begin{subfigure}{0.49\linewidth}
        \centering
        \includegraphics[width=0.7\linewidth]{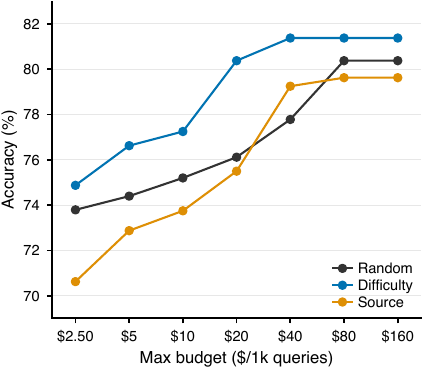}
        \caption{20-model set: uniform costs.}
    \end{subfigure}
    \begin{subfigure}{0.49\linewidth}
        \centering
        \includegraphics[width=0.7\linewidth]{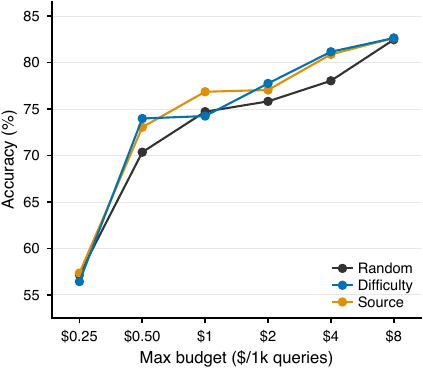}
        \caption{RouterBench: realized costs.}
    \end{subfigure}
    \begin{subfigure}{0.49\linewidth}
        \centering
        \includegraphics[width=0.7\linewidth]{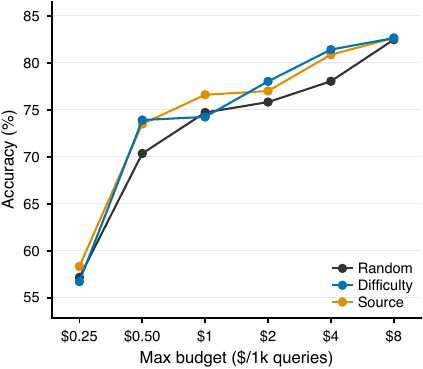}
        \caption{RouterBench: uniform costs.}
    \end{subfigure}
    \begin{subfigure}{0.49\linewidth}
        \centering
        \includegraphics[width=0.7\linewidth]{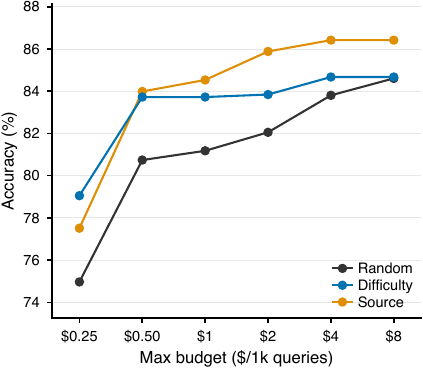}
        \caption{SPROUT: realized costs.}
    \end{subfigure}
    \begin{subfigure}{0.49\linewidth}
        \centering
        \includegraphics[width=0.7\linewidth]{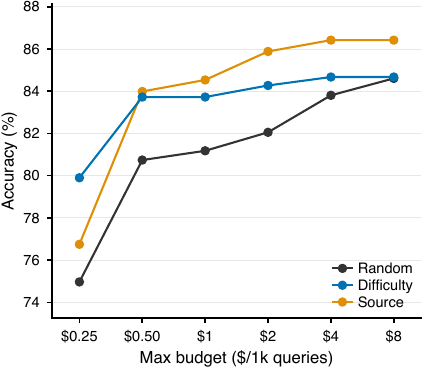}
        \caption{SPROUT: uniform costs.}
    \end{subfigure}
    \begin{subfigure}{0.49\linewidth}
        \centering
        \includegraphics[width=0.7\linewidth]{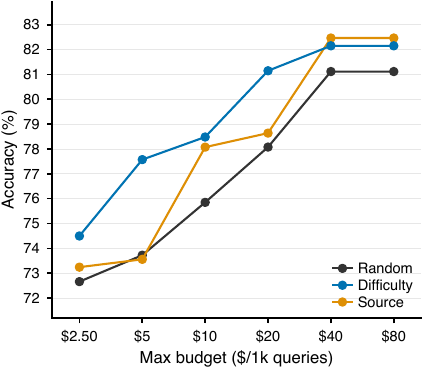}
        \caption{LLMRouterBench: realized costs.}
    \end{subfigure}
    \begin{subfigure}{0.49\linewidth}
        \centering
        \includegraphics[width=0.7\linewidth]{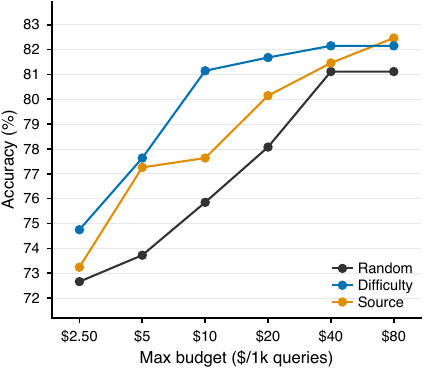}
    \caption{LLMRouterBench: uniform costs.}
    \end{subfigure}
    \caption{Replication of Figure~\ref{fig:semantic} for all other datasets. For the three benchmarks, we split the dataset into training and evaluation halves. 95\% half-widths do not exceed 3.2~pp (20-model set), 0.7~pp (RouterBench), 0.4~pp (SPROUT), and 2.1~pp (LLMRouterBench). On the 20-model set, difficulty routing underperforms random at \$2.50/1k queries because many of the most difficult queries also have the longest outputs, and this pattern reverses using uniform costs, an example of accuracy hacking. On SPROUT, source-based routing outperforms difficulty-based routing, suggesting that its model set exhibits some noticeable specialization.}
    \label{fig:semantic-all}
\end{figure*}

\begin{figure*}
    \centering
    \includegraphics[width=0.9\linewidth]{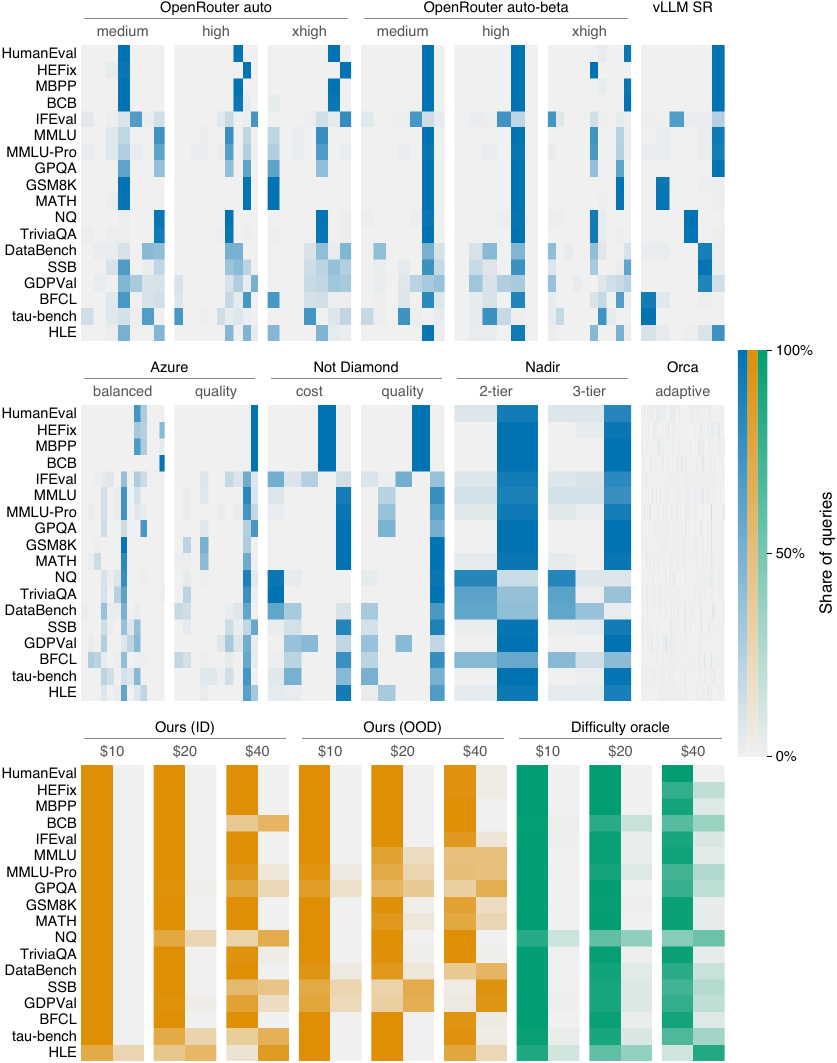}
    \caption{Heatmaps of each router's choices over each data source. We compare all commercial router tiers, our routers at three price points ID and OOD, and a difficulty oracle using two models at the same price points. Models are ranked in increasing order of ability $a_i$. Note that router choices can only be meaningfully compared over identical rosters.}
    \label{fig:source-map}
\end{figure*}

\FloatBarrier
\clearpage
\section{Model Ability Scores}
\label{app:aaii}
In Tables~\ref{tab:router-patterns} and~\ref{tab:ours-id}, a model's ability $a$ is its score on the Artificial Analysis Intelligence Index (AAII) version 4.1.1. Since $\gamma$ only depends on how models are ordered, a small error in one score only matters if it changes the order of two models that a router actually uses.

For every chosen model, we find the AAII listing that best matches the reasoning effort. Table~\ref{tab:aaii-roster} lists the 26 models from our model set. As discussed in Appendix~\ref{app:setup}, we store each commercial router's chosen model on each query and call the model on OpenRouter using our generation configuration to ensure fairness. For some models, we send no effort setting and get the provider's default, and we try to infer the default from the model's output length. For example, GPT-5.5 with no setting writes about as much as when we ask for medium effort (1{,}323 vs 1{,}311 tokens on average), so we give it the medium score. Table~\ref{tab:aaii-commercial} lists the other 76 models that commercial routers picked. For these, we do not have output-length evidence, so when the model ID does not name an effort, we use the default effort listed on AA.

We match AAII scores for each model using the first available option:
\begin{enumerate}
    \item \textbf{Exact:} AA scores this model at the same effort.
    \item \textbf{Only published effort:} AA only includes one score for the model across effort levels, so we use that score.
    \item \textbf{Default effort:} The model ID does not name an effort, so we use the default AA effort.
    \item \textbf{Interpolated:} AA does not list the model at our chosen effort level, so we take the midpoint of the effort levels just above and below it.
    \item \textbf{Converted from v4.3.2:} AA only provides v4.3.2 scores, and we convert to a v4.1.1 score using $\text{v4.1.1} = -2.21 + 1.72 \cdot \text{v4.3.2}$, which we fit on 26 models that both versions score.
    \item \textbf{Converted from v4.1:} We follow the same procedure as for v4.3.2, converting via $\text{v4.1.1} = \text{v4.1} + 1.00$, which is the average gap across nine models that both versions score.
    \item \textbf{Stand-in:} AA does not score the model, so we use the closest model with a score.
    \item \textbf{Rank only:} Only applies to Mistral NeMo, which has no AA score in any version. Our capability estimates for it indicate that it is the weakest model chosen by Not Diamond, so we give it a score of 4, below that of every other model.
\end{enumerate}

Although the AAII scores are not exact, the imperfect matches cannot explain the gap in $\gamma(d,a)$ between commercial routers and our router (Table~\ref{tab:router-patterns}). Of the routed queries, 55\% go to a model with an exact AA match and another 31\% go to a model scored at AA's default or only published effort. To test how much these scores matter, we perturb every inexactly matched model's score by an independent random amount of up to $\pm5$ points and recompute $\gamma$ over 1{,}000 draws. Across all routers, the reported values span $-0.16$ to $+0.18$, and across all 1{,}000 draws they span $-0.19$ to $+0.19$. Even under the most favorable ordering of each router's models, no commercial router exceeds $\gamma = +0.27$, well below the $+0.40$ of our router OOD and $+0.81$ ID at \$10/1k queries.

\onecolumn
{\small
\setlength{\tabcolsep}{4pt}
\renewcommand{\arraystretch}{1.1}
\setlength{\LTcapwidth}{\textwidth}
\begin{longtable}{P{0.21\linewidth} P{0.19\linewidth} P{0.24\linewidth} S[table-format=2.1] P{0.19\linewidth}}
\caption{AAII scores for the 26 models in our model set, sorted by AAII. \emph{Effort} is the reasoning setting we sent, \emph{AA entry used} is the matched Artificial Analysis listing, and \emph{Match} is the matching method.}
\label{tab:aaii-roster} \\
\toprule
Model & Effort & AA entry used & {AAII} & Match \\
\midrule
\endfirsthead
\multicolumn{5}{l}{\textit{\tablename~\thetable, continued}} \\
\toprule
Model & Effort & AA entry used & {AAII} & Match \\
\midrule
\endhead
\bottomrule
\multicolumn{5}{r}{\textit{Continued on next page}} \\
\endfoot
\bottomrule
\endlastfoot
Claude Opus 5 (high) & high & Claude Opus 5 (high) & 61 & Exact \\
Grok 4.6 (high) & high & Grok 4.6 (high) & 61 & Exact \\
GPT 5.6 Sol (xhigh) & xhigh & GPT-5.6 Sol (xhigh) & 59 & Exact \\
Claude Opus 4.8 & high & Claude Opus 4.8 (max) & 57 & Only published effort \\
Gemini 3.7 Flash & none & Gemini 3.7 Flash (high) & 56 & Default effort \\
GLM-5.2 & none (behaves like high) & GLM-5.2 (max) & 53 & Only published effort \\
GLM-5.2 (high) & high & GLM-5.2 (max) & 53 & Only published effort \\
DeepSeek V4 Flash 0731 (thinking) & 4k-token budget & DeepSeek V4 Flash 0731 (max) & 52 & Only published effort \\
GPT 5.5 & none (defaults to medium) & GPT-5.5 (medium) & 51 & Exact \\
GPT 5.5 (medium) & medium & GPT-5.5 (medium) & 51 & Exact \\
Gemini 3.1 Pro & none & Gemini 3.1 Pro Preview & 48 & Exact \\
GPT 5.4 & none (defaults to medium) & GPT-5.4 (medium) & 47.8 & Interpolated between low and xhigh \\
GPT 5.6 Luna (high) & high & GPT-5.6 Luna (high) & 47 & Exact \\
DeepSeek V4 Pro & none (reasons anyway) & DeepSeek V4 Pro (Reasoning, High) & 44.1 & Converted from v4.1 (43.1) \\
DeepSeek V4 Pro (thinking) & 4k-token budget & DeepSeek V4 Pro (Reasoning, High) & 44.1 & Converted from v4.1 (43.1) \\
GLM-5.1 & none & GLM-5.1 & 41 & Exact \\
DeepSeek V4 Flash & none (reasons anyway) & DeepSeek V4 Flash (Reasoning, High) & 38.5 & Converted from v4.1 (37.5) \\
DeepSeek V4 Flash (thinking) & 4k-token budget & DeepSeek V4 Flash (Reasoning, High) & 38.5 & Converted from v4.1 (37.5) \\
Claude Sonnet 4.6 & no thinking & Claude Sonnet 4.6 (Non-reasoning, high) & 37 & Exact \\
GPT 5.4 Mini & none (defaults to medium) & GPT-5.4 mini (medium) & 30.8 & Converted from v4.1 (29.8) \\
Gemini 3.1 Flash (low) & low & Gemini 3.1 Flash Lite Preview & 26 & Only published effort \\
Gemini 3.1 Flash & none & Gemini 3.1 Flash Lite Preview & 26 & Exact \\
DeepSeek V3 & none & DeepSeek V3.2 (Non-reasoning) & 25.4 & Converted from v4.3.2 (16) \\
Claude Haiku 4.5 & no thinking & Claude 4.5 Haiku (Non-reasoning) & 24 & Exact \\
gpt-oss-120b & none (defaults to medium) & gpt-oss-120b (medium) & 19.5 & Interpolated between low and high \\
gpt-oss-20b (low) & low & gpt-oss-20b (low) & 15 & Converted from v4.3.2 (10) \\
\end{longtable}}

{\small
\setlength{\tabcolsep}{4pt}
\renewcommand{\arraystretch}{1.1}
\setlength{\LTcapwidth}{\textwidth}
\begin{longtable}{P{0.27\linewidth} S[table-format=4.0, group-minimum-digits=4, group-separator={,}] P{0.25\linewidth} S[table-format=2.1] P{0.24\linewidth}}
\caption{AAII scores for each model outside our model set chosen by a commercial router, sorted by AAII. \emph{Model ID} is the ID returned by the router, \emph{Queries} is the number of queries the model answered across all commercial routers, \emph{AA entry used} is the matched Artificial Analysis listing, and \emph{Match} is the matching method.}
\label{tab:aaii-commercial} \\
\toprule
Model ID & {Queries} & AA entry used & {AAII} & Match \\
\midrule
\endfirsthead
\multicolumn{5}{l}{\textit{\tablename~\thetable, continued}} \\
\toprule
Model ID & {Queries} & AA entry used & {AAII} & Match \\
\midrule
\endhead
\bottomrule
\multicolumn{5}{r}{\textit{Continued on next page}} \\
\endfoot
\bottomrule
\endlastfoot
claude-fable-5 & 1 & Claude Fable 5 & 62 & Default effort \\
glm-5.3 & 5 & GLM-5.3 (max) & 60 & Default effort \\
kimi-k3 & 735 & Kimi K3 (max) & 60 & Default effort \\
qwen3.8-max & 1 & Qwen3.8 Max & 58 & Exact \\
gpt-5.6-sol-high & 1746 & GPT-5.6 Sol (high) & 57 & Exact \\
muse-spark-1.2 & 3 & Muse Spark 1.2 (xhigh) & 57 & Default effort \\
gpt-5.5-pro & 10 & GPT-5.5 (xhigh) & 56 & Stand-in: the non-Pro version, a lower bound \\
grok-4.5 & 202 & Grok 4.5 (high) & 56 & Default effort \\
claude-opus-4-7 & 7 & Claude Opus 4.7 (max) & 55 & Default effort \\
claude-sonnet-5 & 451 & Claude Sonnet 5 (max) & 55 & Only published effort \\
deepseek-v4-pro-0813 & 16 & DeepSeek V4 Pro 0813 (max) & 53 & Exact \\
muse-spark-1.1 & 6 & Muse Spark 1.1 (xhigh) & 53 & Default effort \\
gpt-5.4-pro & 10 & GPT-5.4 (xhigh) & 52.4 & Stand-in: the non-Pro version, a lower bound, converted from v4.1 (51.4) \\
deepseek-v4-flash-vision-exp & 8 & DeepSeek V4 Flash 0731 (max) & 52 & Stand-in: the text model it matches \\
gemini-3.6-flash & 6 & Gemini 3.6 Flash & 52 & Exact \\
qwen3.8-27b & 28 & Qwen3.8 27B (xhigh) & 52 & Default effort \\
gemini-3.1-pro-preview-customtools & 1 & Gemini 3.1 Pro Preview & 48 & Stand-in: the same model without tool tuning \\
gemini-pro-latest & 7 & Gemini 3.1 Pro Preview & 48 & Exact \\
gemini-3.5-flash & 18 & Gemini 3.5 Flash (medium) & 47 & Only published effort \\
gpt-5.6-terra-medium & 136 & GPT-5.6 Terra (medium) & 47 & Exact \\
qwen3.7-max & 11 & Qwen3.7 Max & 47 & Exact \\
gpt-5.3-codex & 10 & GPT-5.3 Codex (xhigh) & 46 & Default effort \\
Kimi-K2.6 & 1 & Kimi K2.6 & 45 & Exact \\
minimax-m3 & 9 & MiniMax-M3 & 45 & Exact \\
qwen3.6-plus & 6 & Qwen3.6 Plus & 44.3 & Converted from v4.3.2 (27) \\
gpt-5.2 & 15 & GPT-5.2 (xhigh) & 43 & Default effort \\
gpt-5.2-pro & 8 & GPT-5.2 (xhigh) & 43 & Stand-in: the non-Pro version, a lower bound \\
kimi-k2.7-code & 7 & Kimi K2.7 Code & 43 & Exact \\
hy3 & 6 & Hy3 & 42 & Exact \\
glm-5 & 2 & GLM-5 & 41 & Exact \\
gpt-5.2-codex & 6 & GPT-5.2 Codex (xhigh) & 41 & Default effort \\
gpt-5.4-nano & 25 & GPT-5.4 nano (xhigh) & 40 & Default effort \\
claude-opus-4-6 & 1 & Claude Opus 4.6 (Non-reasoning, high) & 39 & Default effort \\
gemini-3-flash-preview & 273 & Gemini 3 Flash (reasoning) & 39 & Stand-in: the released version \\
minimax-m2.7 & 14 & MiniMax-M2.7 & 39 & Exact \\
qwen3.7-plus & 4 & Qwen3.7 Plus & 39 & Exact \\
grok-4.20 & 369 & Grok 4.20 0309 v2 & 38 & Exact \\
grok-4.3 & 8 & Grok 4.3 (medium) & 38 & Default effort \\
gemini-3.5-flash-lite & 5 & Gemini 3.5 Flash-Lite & 37 & Exact \\
gpt-5.1 & 8 & GPT-5.1 (high) & 37 & Default effort \\
claude-opus-4-5 & 4 & Claude Opus 4.5 (Non-reasoning) & 36 & Default effort \\
gpt-5.1-codex & 7 & GPT-5.1 Codex (high) & 36 & Default effort \\
glm-4.7 & 7 & GLM 4.7 & 35 & Exact \\
gpt-5 & 100 & GPT-5 (high) & 35 & Default effort \\
gpt-5-pro & 8 & GPT-5 (high) & 35 & Stand-in: the non-Pro version, a lower bound \\
qwen3.5-27b & 5 & Qwen3.5-27B & 35 & Exact \\
minimax-m2.5 & 15 & MiniMax-M2.5 & 34 & Exact \\
qwen3.5-397b-a17b & 7 & Qwen3.5 397B A17B & 34 & Exact \\
qwen3.5-plus-20260420 & 6 & Qwen3.5 397B A17B & 34 & Stand-in: the open model behind the hosted API \\
qwen3.5-122b-a10b & 8 & Qwen3.5 122B A10B & 33 & Exact \\
qwen3.6-35b-a3b & 13 & Qwen3.6 35B A3B & 32 & Exact \\
qwen3.6-flash & 20 & Qwen3.6 35B A3B & 32 & Stand-in: the open model behind the hosted API \\
gpt-5.1-codex-mini & 6 & GPT-5.1-Codex-Mini & 31 & Exact \\
claude-sonnet-4.5 & 40 & Claude 4.5 Sonnet (Non-reasoning) & 30.5 & Converted from v4.3.2 (19) \\
qwen3.5-35b-a3b & 12 & Qwen3.5 35B A3B (Reasoning) & 30.5 & Converted from v4.3.2 (19) \\
gemma-4-31B-it & 5 & Gemma 4 31B & 30 & Exact \\
gemini-2.5-pro & 4 & Gemini 2.5 Pro & 26 & Exact \\
gemini-3.1-flash-lite & 4 & Gemini 3.1 Flash Lite Preview & 26 & Stand-in: the preview version \\
gemma-4-26b-a4b-it & 18 & Gemma 4 26B A4B & 26 & Exact \\
gpt-5-mini & 101 & GPT-5 mini (high) & 26 & Default effort \\
qwen3-max & 6 & Qwen3 Max & 24 & Exact \\
qwen3.5-flash-02-23 & 10 & Qwen3.5 35B A3B (Non-reasoning) & 23.6 & Stand-in: non-reasoning variant, converted from v4.3.2 (15) \\
glm-4.6 & 4 & GLM 4.6 & 23 & Exact \\
qwen3-vl-235b-a22b-thinking & 10 & Qwen3 VL 235B A22B (Reasoning) & 20.2 & Converted from v4.3.2 (13) \\
gpt-4.1 & 6 & GPT-4.1 & 20 & Exact \\
gpt-5-nano & 27 & GPT-5 nano (high) & 20 & Default effort \\
gpt-5-search-api & 5 & GPT-5 (minimal) & 16.7 & Stand-in: GPT-5 without reasoning, converted from v4.3.2 (11) \\
gpt-4.1-mini & 15 & GPT-4.1 Mini & 15 & Exact \\
gemini-2.5-flash & 24 & Gemini 2.5 Flash & 14 & Exact \\
qwen3-vl-235b-a22b-instruct & 9 & Qwen3 VL 235B A22B Instruct & 14 & Exact \\
gpt-4o-2024-08-06 & 6 & GPT-4o (Aug '24) & 11.6 & Converted from v4.3.2 (8) \\
qwen3-vl-8b-thinking & 6 & Qwen3 VL 8B (Reasoning) & 11.6 & Converted from v4.3.2 (8) \\
gpt-4o & 4 & GPT-4o (Nov) & 11 & Exact \\
gpt-4o-2024-11-20 & 10 & GPT-4o (Nov) & 11 & Exact \\
gpt-4.1-nano & 21 & GPT-4.1 Nano & 10 & Exact \\
sonar & 150 & Sonar & 9 & Exact \\
gpt-4-turbo & 3 & GPT-4 Turbo & 8 & Exact \\
qwen3-vl-8b-instruct & 6 & Qwen3 VL 8B Instruct & 8 & Exact \\
gemini-2.5-flash-lite & 21 & Gemini 2.5 Flash Lite & 7 & Exact \\
gpt-4o-mini & 8 & GPT-4o-mini & 7 & Exact \\
gpt-3.5-turbo & 7 & GPT-3.5 Turbo & 6.4 & Converted from v4.3.2 (5) \\
gpt-3.5-turbo-1106 & 2 & GPT-3.5 Turbo & 6.4 & Converted from v4.3.2 (5) \\
gpt-3.5-turbo-16k & 6 & GPT-3.5 Turbo & 6.4 & Converted from v4.3.2 (5) \\
open-mistral-nemo & 182 & none & 4 & Rank only: below every other Not Diamond pick \\
\end{longtable}}

\end{document}